\documentclass[nohyperref]{article}

\usepackage{microtype}
\usepackage{graphicx}
\usepackage{subfigure}
\usepackage{booktabs} 
\usepackage{hyperref}

\usepackage[accepted]{icml2022}

\usepackage{amsmath}
\usepackage{amssymb}
\usepackage{mathtools}
\usepackage{amsthm}
\usepackage{xcolor}
\usepackage{enumitem}
\usepackage{xspace}
\usepackage{multirow}
\usepackage{dsfont}

\usepackage[capitalize,noabbrev]{cleveref}

\theoremstyle{plain}
\newtheorem{theorem}{Theorem}[section]

\newtheorem{lemma}[theorem]{Lemma}

\theoremstyle{definition}

\newtheorem{assumption}[theorem]{Assumption}
\theoremstyle{remark}
\newtheorem{remark}[theorem]{Remark}

\usepackage[textsize=tiny]{todonotes}
\usepackage{Definitions}
\allowdisplaybreaks

\newcommand{\algabb}{{CTRL}\xspace}
\newcommand{\ucblinear}{{CTRL-UCBL}\xspace}
\newcommand{\ucbmlp}{{CTRL-UCBM}\xspace}

\newenvironment{proof_sketch}{%
\proof}{\endproof}

\icmltitlerunning{Making Linear MDPs Practical via Contrastive Representation Learning}

\begin{document}

\twocolumn[
\icmltitle{Making Linear MDPs Practical via Contrastive Representation Learning}



\icmlsetsymbol{equal}{*}

\begin{icmlauthorlist}
\icmlauthor{Tianjun Zhang}{equal,berkeley}
\icmlauthor{Tongzheng Ren}{equal,texas,brain}
\icmlauthor{Mengjiao Yang}{berkeley,brain}
\icmlauthor{Joseph E. Gonzalez}{berkeley}
\icmlauthor{Dale Schuurmans}{brain,alberta}
\icmlauthor{Bo Dai}{brain}
\end{icmlauthorlist}

\icmlaffiliation{berkeley}{UC Berkeley}
\icmlaffiliation{texas}{UT Austin}
\icmlaffiliation{brain}{Google Brain}
\icmlaffiliation{alberta}{University of Alberta}

\icmlcorrespondingauthor{Tianjun Zhang}{tianjunz@berkeley.edu}
\icmlcorrespondingauthor{Tongzheng Ren}{tongzheng@utexas.edu}
\icmlcorrespondingauthor{Bo Dai}{bodai@google.com}
\icmlkeywords{Machine Learning, ICML}

\vskip 0.3in
]



\printAffiliationsAndNotice{\icmlEqualContribution} 

\begin{abstract}

It is common to address the curse of dimensionality in Markov decision processes (MDPs) by exploiting low-rank representations. This motivates much of the recent theoretical study on linear MDPs. However, most approaches require a given representation under unrealistic assumptions about the normalization of the decomposition or introduce unresolved computational challenges in practice.
Instead, we consider an alternative definition of linear MDPs that automatically ensures normalization while allowing efficient representation learning via contrastive estimation. The framework also admits confidence-adjusted index algorithms, enabling an efficient and principled approach to incorporating optimism or pessimism in the face of uncertainty.
To the best of our knowledge, this provides the first practical representation learning method for linear MDPs that achieves both strong theoretical guarantees and empirical performance.  \emph{Theoretically}, we prove that the proposed algorithm is sample efficient in both the online and offline settings.  \emph{Empirically}, we demonstrate superior performance over existing state-of-the-art model-based and model-free algorithms on several benchmarks. 

\end{abstract}

\setlength{\abovedisplayskip}{2pt}
\setlength{\abovedisplayshortskip}{2pt}
\setlength{\belowdisplayskip}{2pt}
\setlength{\belowdisplayshortskip}{2pt}
\setlength{\jot}{2pt}

\setlength{\floatsep}{2ex}
\setlength{\textfloatsep}{2ex}


\section{Introduction}\label{sec:intro}

In reinforcement learning~(RL), the goal of an agent is to learn a policy that maximizes the expected reward accumulated during sequential interaction with an unknown environment. Most RL algorithms are developed under the framework of Markov decision processes~(MDPs)~\citep{puterman2014markov}, where performance can be evaluated by ``regret'', \ie, the difference in accumulated reward between the best policy and the policies produced by the algorithm. It is known that in the tabular case where the number of states $\abr{\Scal}$ and actions $\abr{\Acal}$ are finite, there exist algorithms that can achieve the asymptotic regret lower bound $\Omega\rbr{\sqrt{\abr{\Scal}\abr{\Acal}}\cdot\sqrt{T}}$~\citep{jaksch2010near,dann2015sample,osband2016lower,azar2017minimax,jin2018q} after $T$ iterations, which is unimprovable in the worst case. Unfortunately, the polynomial dependence on $\abr{\Scal}$ and $\abr{\Acal}$ impedes the application of the regret bounded RL algorithms in practice, where the number of states and actions can be very large or even infinite. This difficulty is known as the ``curse of dimensionality'' in control and RL~\citep{bellman1966dynamic}. 

To tackle the curse of dimensionality, low-rank representations via function approximation have been used to exploit additional structures in an MDP. One prominent direction is the linear MDP framework~\citep{yang2019sample, yang2020reinforcement,jin2020provably}, which leverages additional assumptions to reduce the algorithmic and theoretical challenge of coping with the dimensionality issue in RL. In particular, assuming a \emph{known} low-rank feature that can represent the transition operator and reward function in a linear MDP, then simple optimistic algorithms can achieve regret that is \emph{independent} of the cardinality of the state and action sets. Therefore, the linear MDP framework partially breaks the curse of dimensionality, provides a deeper understanding of the empirical success of deep RL~\citep{mnih2015human,levine2016end}, and can deliver performance improvements. However, the requirement that the low-rank features be known beforehand is difficult to achieve in practice, which has limited the practical impact of linear MDPs in real applications.

Recently, there have been important new progress in making linear MDPs more practical by introducing representation learning techniques that satisfy provable guarantees. ~\citet{agarwal2020flambe} proposed the first theoretically-grounded representation learning algorithm for linear MDPs, which achieves polynomial sample complexity given a maximum likelihood estimation~(MLE) oracle for stochastic transition dynamics. Later, \citet{uehara2021representation}  improved their algorithm by exploiting upper confidence bounds~(UCB) to reduce sample complexity under the same MLE oracle. These efforts have revealed the possibility of achieving statistically efficient representation learning in linear MDPs. However, these algorithms have not addressed the underlying computational issues, leaving a gap between the theory and practice of linear MDPs. Specifically, the transition dynamics in the definition of a linear MDP do not necessarily form a density model, meaning that there might not be a computationally efficient way to implement an MLE oracle. Moreover, the classical formulation of a linear MDP requires an extra normalization condition on the representation, which is also computationally challenging. 

\paragraph{Our contributions.} In this paper, we address the lingering computational efficiency issues in representation learning for linear MDPs. We provide solutions to the aforementioned difficulties, bridge the gap between theory and practice, and eventually establish a sound yet practical foundation for learning in linear MDPs. More specifically:
\begin{itemize}[leftmargin=14pt,topsep=0pt,parsep=0pt,partopsep=0pt,itemsep=-1pt]
    \item We clarify the importance of the normalization condition and propose a variant of linear MDPs wherein it is easy to enforce normalization (\secref{sec:mod_linear_mdp});
    \item We develop the \emph{ConTrastive Representation Learning algorithm,~\algabb} that can implicitly satisfy the density requirement (\secref{sec:nce_mle}); 
    \item We incorporate the representation learning techniques in optimistic and pessimistic confidence-adjusted RL algorithms, \emph{\algabb-UCB} and \emph{\algabb-LCB}, for both online and offline RL problems (\secref{sec:full_alg}), and establish their sample complexities  respectively (\secref{sec:justification}); 
    \item We conduct a comprehensive comparison to  existing state-of-the-art model-based and model-free RL algorithms on several benchmarks for the online and offline settings, demonstrating superior empirical performance of the proposed \algabb (\secref{sec:exp}).
\end{itemize}

\subsection{Related Work}\label{sec:related_work}

There have been several attempts to achieve provably sound and effective representation learning within more restricted models that make extra assumptions.
 For example, \citet{du2019provably,misra2020kinematic} considered  representation learning in \emph{block MDPs};~\citet{modi2021model} proposed a model-free representation learning algorithm for \emph{low-nonnegative rank MDPs}. These are special case of linear MDPs and are quite limited in terms of expressiveness, as shown in~\citep{agarwal2020flambe}. We emphasize that these algorithms also  primarily focus on sample complexity, and do not necessarily address all the computational complexity challenges. 

One exception is \citet{ren2021free}, which investigates an alternative proposal for exploiting  linear MDP structure to achieve provable algorithms in practical problems based on
a different solution to the computational intractability issue.  Their proposal is to maintain a \emph{posterior} of the control model under specific \emph{state-independent noise}, \eg, Gaussian noise, 
exploiting the explicit Fourier transform of the Gaussian distribution to recover an equivalent \emph{infinite-dimensional} linear MDP posterior.
By contrast, the method proposed here only requires a \emph{point estimate} of the features. Moreover, we do not restrict the dynamics and noise model, attaining greater \emph{generality}, while learning a \emph{low-rank} feature representation that is more compact and effective than a random features approximation. 

Beyond theoretically-justified representation learning techniques for RL, many representation learning methods have been proposed for different purposes, such as: reconstruction~\citep{watter2015embed,li2019learning,hafner2019learning},  bi-simulation~\citep{ferns2004metrics,gelada2019deepmdp,castro2020scalable,zhang2020learning}, invariance preservation~\citep{zhang2020invariant}, successor predictive occurrence backup~\citep{dayan1993improving,barreto2016successor,kulkarni2016deep}, and spectral decomposition of various state-state transition operators~\citep{mahadevan2007proto,wu2018laplacian,duan2018state}. Contrastive losses have also been introduced for representation learning in RL. For instance,~\citep{yang2020plan2vec} extends Contrastive Predictive Coding~\citep{oord2018representation} for planning with specially designed positive pairs for \emph{state} representation learning;~\citep{nachum2021provable,yang2021trail} considered a contrastive loss for training an energy-based model with offline data to produce an \emph{infinite-dimensional} representation for  imitation learning in \emph{finite-action MDPs}. 
Note that, beyond the various differences between these approaches to representation learning (state vs.\ state-action modeling, imitation learning vs.\ policy optimization, finite vs.\ continuous actions), a major differentiator of this paper is that \algabb is naturally developed with UCB or LCB to support exploration or conservative exploitation respectively, while previous work considered passive data collection that was \emph{independent} of such optimistic or pessimistic considerations. 
This property of our proposed method allows provable sample complexity bounds for both online and offline settings. 


\section{Preliminaries}\label{sec:prelim}

We first provide a brief introduction to linear MDPs and noise contrastive estimation, which will play an important role in the algorithm design. 

\subsection{Markov Decision Processes (MDPs)}
Markov Decision Processes (MDPs) are a standard model of sequential decision making considered in reinforcement learning.  An MDP is specified by a tuple $\mathcal{M} = (\mathcal{S}, \mathcal{A}, r, P, \rho, \gamma)$, where $\mathcal{S}$ is the state space, $\mathcal{A}$ is the action space, $r:\mathcal{S} \times \mathcal{A} \to [0, 1]$ is the reward function, $P:\mathcal{S}\times \mathcal{A} \to \Delta(\mathcal{S})$ is the transition operator, $\rho$ is an initial state distribution and $\gamma\in (0, 1]$ is the discount factor. A policy $\pi$ can be defined as $\pi:\mathcal{S}\to\Delta(\mathcal{A})$ where $\Delta\rbr{\Acal}$ denotes the distributions over domain $\Acal$. Following the standard notation, we define the value function $V_{P, r}^\pi(s) := \mathbb{E}_{P, \pi}\left[\sum_{t=0}^{\infty} \gamma^t r(s_t, a_t)|s_0 = s\right]$ and the action-value function $Q_{P, r}^\pi(s, a) = \mathbb{E}_{P, \pi}\left[\sum_{t=0}^{\infty} \gamma^t r(s_t, a_t)|s_0 = s, a_0 = a\right]$, which are the expected discounted cumulative rewards under transition $P$ and reward $r$ when executing policy $\pi$. From these two definitions, it is straightforward to establish the following recursive relationship:
\begin{align}
    Q_{P, r}^\pi(s, a) = & r(s, a) + \gamma\mathbb{E}_{s^\prime\sim P(\cdot|s, a)} \sbr{V_{P, r}^\pi(s^\prime)},\label{eq:bellman_q}\\
    V_{P, r}^\pi(s) = & \mathbb{E}_{a\sim \pi(s)} \sbr{Q_{P, r}^\pi(s, a)}.
\end{align}
For simplicity, with slightly abuse of notation, we define $V_{P, r}^\pi = \mathbb{E}_{s\sim \rho} \left[V_{P, r}^{\pi}(s)\right]$. Furthermore, we additionally define the occupancy measure:
\begin{align}
\textstyle
d_{P}^\pi(s, a) = (1-\gamma)^{-1}\mathbb{E}_{s_0\sim\rho, a_0\sim\pi(s_0)}\left[\sum_{t=0}^\infty \gamma^t \mathbf{1}(s, a)\right],
\end{align}
where $\mathbf{1}(\cdot)$ is the indicator function. Clearly, $d_{P}^\pi$ is a probability measure on $(s, a)$ that satisfies
\begin{align}
    \mathbb{E}_{(s, a)\sim\rho\times \pi}\left[Q_{P, r}^\pi(s, a)\right] = \mathbb{E}_{(s, a) \sim d_{P}^\pi(s, a)} \left[r(s, a)\right],
\end{align}
as shown in~\citet{nachum2020reinforcement}. 

When $|\mathcal{S}|$ and $|\mathcal{A}|$ are large or even infinite, it is necessary to incorporate  function approximation to exploit MDP structure, to address the curse of dimensionality. One of the most prominent forms of structured MDP that allows for simple function approximation is linear MDPs, which add the following structural assumption:
\begin{assumption}[Linear MDP \citep{jin2020provably}]
\label{assump:linear_mdp}
An MDP $\mathcal{M}=(\mathcal{S}, \mathcal{A}, H, P, r)$ is called a \emph{linear MDP} if there exists two feature maps $\phi:\mathcal{S}\times \mathcal{A} \to \mathbb{R}^d$, $\mu:\mathcal{S} \to \mathbb{R}^d$ and a vector $\theta_r \in \mathbb{R}^d$, such that
\begin{align}
    P(s^\prime|s, a) = \langle \phi(s, a), \mu(s^\prime)\rangle, \quad r(s, a) = \langle \phi(s, a), \theta_r\rangle
    \label{eq:linear_mdp_def}
\end{align}
and $\phi$ and $\mu$ also satisfy the normalization conditions: 
\begin{align}
    & \forall (s, a),~ \|\phi(s, a)\|_2 \leq 1, ~\|\theta_r\|_2 \leq \sqrt{d}\\
    & \forall g:\mathcal{S}\to\mathbb{R},~\|g\|_{L_{\infty}}\leq 1~ \Rightarrow\left\|\int_{\mathcal{S}}\mu(s^\prime)g(s^\prime) d s^\prime\right\|_2 \leq \sqrt{d} \label{eq:normalization_condition}
\end{align}
\end{assumption}
An essential property of a linear MDP is that $Q_{P, r}^\pi(s, a)$ for \emph{any} policy $\pi$ is linear w.r.t.\ $\phi(s, a)$ \citep{jin2020provably}; that is, when \eqref{eq:linear_mdp_def} is applied to the right hand size of \eqref{eq:bellman_q} we obtain:
\begin{align*}
    Q_{P, r}^\pi(s, a) = & r(s, a) + \gamma\int V_{P, r}^\pi(s) P(s^\prime|s, a) d s^\prime \\
    & = \left\langle \phi(s, a), \theta_r 
    + \gamma\int V_{P, r}^\pi(s^\prime) \mu(s^\prime) d s^\prime\right\rangle.
\end{align*}
Consequently, one can naturally incorporate linear function approximation when learning a $Q$ function in a linear MDP, simplifying practical application.

We emphasize that, although the MDP is referred to as ``linear'' here, the model components $\phi\rbr{s, a}$ and $\mu\rbr{s'}$ can be flexibly parametrized by deep models, and thus can be viewed as energy-based models~\citep{lecun2006tutorial} with special potential functions, \ie, $\log \rbr{\langle \phi(s, a), \mu(s^\prime)\rangle}$. 


\subsection{Noise Contrastive Estimation}
Noise contrastive estimation (NCE) \citep{gutmann2010noise, gutmann2012noise} was first proposed to estimate  unnormalized statistical models without requiring explicit calculation of the partition function. For an (unnormalized) statistical model $p_f(x) = \frac{f\rbr{x}}{Z_f}$ (where $f\rbr{x} > 0$ for all $x$ in the support and $Z_f = \int f(x) dx$), maximum likelihood estimation (MLE) is generally intractable due to the presence of $Z_f$. NCE was developed to estimate $f$ from data $\{(x_i)\}_{i\in [n]}$ by the following M-estimator:
\begin{align}
    \textstyle
    \left(\widehat{f}_n^{\mathrm{NCE}}, \widehat{\gamma}_n^{\mathrm{NCE}}\right) =& \mathop{\arg\max}_{f\in\mathcal{\mathcal{F}}, \gamma\in\mathbb{R}} \frac{1}{n}  \sum_{i=1}^n\bigg[ \log(h(x_i;f, \gamma)) \nonumber\\
     & + \textstyle \sum_{j=1}^K \log(1-h(y_{i, j}; f, \gamma)) \bigg],
     \label{eq:binary}
\end{align}
where $\{y_{i, j}\}_{i\in [n], j\in [K]}$ are artificially generated data sampled from a predefined noise distribution $q(y)$ that has full support on the domain of interest, and $h(x; f, \gamma)$ is defined by
\begin{align}
    h(x;f, \gamma) = \frac{f(x)\exp(-\gamma)/q(x)}{f(x)\exp(-\gamma)/q(x) + K}.
\end{align}
Here $\widehat{\gamma}_n^{\mathrm{NCE}}$ is an estimator of the normalizing constant $Z_f$. The objective function~\eqref{eq:binary} is also known as the binary-based objective.  It can be implemented via a logistic regression that distinguishes real data from generated noise. 

The binary objective works well for an unconditional model. However, as \citet{ma2018noise} point out, when \eqref{eq:binary} is used to estimate a \emph{conditional} model $p(x|u) = \frac{f\rbr{x, u}}{Z_f\rbr{u}}$ (where $Z_f\rbr{u} = \int f\rbr{x, u}dx$), NCE cannot provide a consistent estimator, due to the different values of $Z_f\rbr{u}$ for different choices of the conditioning variable value $u$. To resolve this issue, a ranking-based objective has been proposed to provide consistent estimation under conditioning:
\begin{equation}
   \widehat{f}_n^{\mathrm{NCE}} = \mathop{\arg\max}_{f\in\mathcal{F}} \frac{1}{n}\sum_{i=1}^n \log \frac{f(x_i, u_i)/q(x_i)}{\frac{f(x_i, u_i)}{q(x_i)} + \sum_{j=1}^K \frac{f(y_{i, j}, u_i)}{q(y_{i, j})}},
    \label{eq:ranking}
\end{equation}
where $\{(x_i, u_i)\}_{i=1}^n$ are the given data. This estimator~\eqref{eq:ranking} can be viewed as a multi-class extension to distinguishing the real data from  generated noise.

Using standard asymptotic M-estimator theory, \citet{gutmann2010noise, gutmann2012noise, riou2018noise, ma2018noise} have provided consistency and asymptotic normality results for the parametric model under the binary-based objective, while \citet{ma2018noise}  provided similar results for the ranking-based objective.

\section{The Difficulties of Practical Linear MDP}\label{sec:hardness}


Here we explicitly identify the computational difficulties in applying linear MDP,  paving the way to making it practical.

\subsection{Normalization Conditions}\label{sec:norm_cond}

The normalization condition in~\asmpref{assump:linear_mdp} might seem strange at first glance. One could ask why there is a constraint $\left\|\int_{\mathcal{S}}\mu(s^\prime) g(s^\prime)d s^\prime\right\|_2\leq\sqrt{d}$ 
(for all $\|g\|_{L_\infty}\leq1$)
instead of any other  function of $d$? 
To motivate this choice, consider
the tabular MDP, where $\abr{\mathcal{S}} < \infty$, $\abr{\mathcal{A}} < \infty$, which can be viewed as a special linear MDP with $d = |\mathcal{S}| |\mathcal{A}|$,  $\phi(s, a) = \mathbf{e}_{(s, a)}$ for $\mathbf{e}$ the canonical basis of $\mathbb{R}^{|\mathcal{S}||\mathcal{A}|}$, and $\mu(s^\prime)_{(s, a)} = P(s^\prime|s, a)$. It is immediate in this case that $\|\phi(s, a)\|_2 = 1$ and $ \int_{\mathcal{S}} \mu(s^\prime) d s^\prime = \one \in \RR^{\abr{\Scal}\abr{\Acal}}$ (where $\one$ denotes the vector with all entries equal to $1$), which automatically satisfies the normalization condition. This constraint plays a critical role in obtaining a polynomial dependence on $d$ in a worst case regret bound in linear MDPs, as it guarantees that \citep[][Lemma B.1]{jin2020provably}
\begin{align*}
    \left\|\int_{\mathcal{S}} V_{P, r}^\pi(s^\prime) \mu(s^\prime) ds^\prime\right\|_2 \leq (1-\gamma)^{-1}\sqrt{d}
\end{align*}
noting that $\|V_{P, r}^\pi\|_{L_{\infty}}\leq (1-\gamma)^{-1}$. Such a guarantee is crucial both for the theoretical analysis and practical implementation, as it ensures the linear coefficients needed to express the $Q$ function are not too large, which guarantees low statistical complexity in estimation.

However, when we want to learn $\phi$ and $\mu$ using a function approximator like a neural network, there is no practical mechanism for guaranteeing or even checking the normalization condition. Note that condition \eqref{eq:normalization_condition} requires exact knowledge of the Lebesgue integral $\int_{\mathcal{S}}\mu(s^\prime) g(s^\prime)d s^\prime$ for an \emph{indefinite} $g(s^\prime)$. Beyond a tabular MDP, even if $g$ were known ahead of time, checking the constraint would be difficult: 
in a discrete space $\mathcal{S}$, the Lebesgue integral involves a summation over the entire set, 
whereas numerical methods would be needed in a continuous $\mathcal{S}$,
 which can be computationally inefficient or even fail in high dimensional spaces. 
More significantly,
guaranteeing \eqref{eq:normalization_condition} requires the solution to an optimization on the functional $g$. Given that evaluating the Lebesgue integral for a given $g$ would already be challenging, there is little hope for solving the optimization of $g$ with a computationally efficient method.

Therefore, this normalization condition can be quite restrictive and  impedes  practical application of linear MDPs. 
Even in cases where the linear factorization condition \eqref{eq:linear_mdp_def} can be satisfied, ensuring the normalization condition \eqref{eq:normalization_condition} remains difficult. For example, \citet{ren2021free} show that for nonlinear control problems with Gaussian perturbation, special properties of the Gaussian kernel can be leveraged to obtain an infinite dimensional linear representation that satisfies \eqref{eq:linear_mdp_def}; however, without further assumptions this does not provide adequate bounds for the term in \eqref{eq:normalization_condition}. Such an observation is disappointing given that \eqref{eq:linear_mdp_def} provides a good criterion for learning the representation.



\subsection{MLE Computation Oracle}\label{sec:mle_comp}
In practice, one only has access to raw observations of state and action, which by themselves do not provide an adequate representation for a linear MDP. To resolve this issue, the seminal work \citep{agarwal2020flambe} proposed to learn the representations $\phi$ and $\mu$ using MLE, but at a cost of a computationally inefficient policy-cover  method to ensure sufficient exploration. Later, \citet{uehara2021representation} replaced policy-cover  exploration with confidence-adjusted index form exploration, which is far more computationally efficient. 
However, this does not completely solve the problem, since MLE for conditional model estimation itself can be computationally intractable, due to the constraints on the marginal regularity, \ie,
\begin{align*}
    \left\langle \phi(s, a), \int_{\mathcal{S}} \mu(s^\prime) ds^\prime\right\rangle = 1,\quad  \forall (s, a)\in\mathcal{S}\times \mathcal{A}.
\end{align*}
It is not clear how to solve the optimization problem under such a hard constraint. One idea might be to normalize $\phi$ via $\frac{\phi(s, a)}{\int_{\mathcal{S}} \phi(s, a) \mu(s^\prime) d s^\prime}$, which would naturally satisfy the marginal regularity condition and allow MLE without additional constraints. However, 
this creates new problems at decision making time.
In particular, given a fixed $(s, a)$, one needs to compute the denominator $\int_{\mathcal{S}} \phi(s, a) \mu(s^\prime) d s^\prime$ to extract the corresponding representation,
which can be intractable when $|\mathcal{S}|$ is large or  infinite. Moreover, since $\int_{\mathcal{S}} \phi(s, a) \mu(s^\prime) d s^\prime$ depends on $(s, a)$, the integral would need to be estimated for several $(s, a)\in \Scal\times \Acal$ to choose an action, making the computational issue even more severe. Such an approach  also causes trouble in the representation learning stage when gradient based optimization is used to solve the MLE, since the gradient of the integral would need to be estimated for each  gradient step. 
Hence, the MLE computation oracle assumed in~\citep{agarwal2020flambe,uehara2021representation} is not generally practical from a computational perspective. 


\section{The Proposed Solutions}\label{sec:method}



In this section, we propose a new approach that addresses the two difficulties revealed for learning linear MDPs. 

\subsection{Modified Linear MDP}\label{sec:mod_linear_mdp}
To ensure the normalization condition in~\asmpref{assump:linear_mdp}, we introduce an auxiliary probability measure $p\rbr{s^\prime}$ in conjunction with $\mu\rbr{s^\prime}$ to obtain a modified linear MDP:
\begin{align}\label{eq:mod_linear_mdp}
    P(s^\prime|s, a) = \langle \phi(s, a), {\color{blue}p(s^\prime)}\mu(s^\prime) \rangle.
\end{align}
Here $p(s^\prime)$ can be any probability measure with full support $\Scal$, which plays a similar role to the base measure in exponential families and generalized energy based models~\citep{arbel2020generalized}.
That is, $p(s^\prime)$ will remain fixed but $\mu\rbr{s^\prime}$ will be learned from data. Note that the introduction of $p\rbr{s^\prime}$ does not compromise 
expressiveness in classically considered settings. Take, for example, a tabular MDP as in Section~\ref{sec:norm_cond}: here we can still use $\phi(s, a) = \mathbf{e}_{s, a}$ and let $p(s^\prime)$ be a uniform distribution on $\mathcal{S}$ and $\mu(s^\prime)_{(s, a)} = |\mathcal{S}| P(s^\prime|s, a)$, which  automatically ensures \eqref{eq:mod_linear_mdp}  for all $(s, a, s^\prime)$.

The key benefit of modifying the linear MDP in this manner is that we can now easily control the normalization condition. 
Specifically, for $\|g\|_{L_{\infty}}\leq 1$, by Jensen's inequality, we have 
\begin{align*}
    \left\|\int_{\mathcal{S}} p(s^\prime) \mu(s)^\prime g(s^\prime) ds^\prime\right\|_2^2
    \leq & \int_{\mathcal{S}} p(s^\prime) \|\mu(s^\prime)g(s^\prime)\|_2^2 d s^\prime\\
    \leq & \int_{\mathcal{S}} p(s^\prime) \|\mu(s^\prime)\|_2^2 d s^\prime, 
\end{align*} 
which can now be easily approximated via Monte Carlo estimation instead of numerical integration. This upper bound provides a straightforward way to control  normalization  by introducing a regularizer $\EE_{p\rbr{s^\prime}}\sbr{\nbr{\mu\rbr{s^\prime}}_2^2}$ into the NCE losses~\eqref{eq:binary} and~\eqref{eq:ranking}. 
Alternatively, one could also ensure boundedness simply by introducing a bounded activation, \eg, $\frac{1}{1 + \exp\rbr{\cdot}}$, $\exp(-\rbr{\cdot}^2)$, or $\tanh\rbr{\cdot}$, at the final layer in a neural network parametrization of $\mu\rbr{\cdot}$. 

\begin{algorithm}[tb] 
\caption{\algabb-UCB: Online Exploration with Representation Learning} \label{alg:online_algorithm}
\begin{algorithmic}[1]
  \State \textbf{Input:} Regularizer $\lambda_n$, parameter $\alpha_n$, Model class $\Mcal=\{(\mu,\phi):\mu\in \Psi,\phi\in \Phi\}$, Iteration $N$, Number of Negative Samples $K$.
  \State Initialize $\pi_0(\cdot\mid s)$ to be uniform; set $\Dcal_0 = \emptyset$
  \For{episode $n=1,\cdots,N$ } 
  \State Collect a transition $(s,a,s')$ where $s\sim d_{P^\star}^{\pi_{n-1}}$, $a\sim U(\Acal)$, $s'\sim P^\star(\cdot | s,a)$. \label{line:data_collection}
  \State $\Dcal_n = \Dcal_{n-1} \cup \{s,a,s'\}$ 
  \State Learn representation $\widehat{\phi}(s, a)$ via~\eqref{eq:ranking}. \label{line:representation_online} 
  \State Update the empirical covariance matrix 
  \begin{equation*}
      \widehat\Sigma_n = \sum_{s,a\in\mathcal{D}_n} \widehat\phi_n(s,a) \widehat\phi_n(s,a)^{\top} + \lambda_n I
  \end{equation*}
  \State Set the exploration bonus:
     \begin{equation*}
      \widehat b_n(s,a)=\min \left(\alpha_n \sqrt{\widehat \phi_n(s,a)^{\top}\widehat \Sigma^{-1}_{n}\widehat \phi_n(s,a)},2\right) %
     \end{equation*}\label{line:bonus}
  \State Update policy \label{line:online_plan}
    $ \pi_n=\argmax_{\pi}V^{\pi}_{\widehat P_n,r+\widehat b_n}$ 
  \EndFor
  \State \textbf{Return } $\pi_1,\cdots,\pi_N$
\end{algorithmic}
\end{algorithm}
\subsection{NCE instead of MLE}\label{sec:nce_mle}
Due to the computational intractability of MLE discussed in~\secref{sec:mle_comp},  we exploit  noise-contrastive estimation (NCE) to learn the representations $\phi$ and $\mu$ in~\eqref{eq:mod_linear_mdp}, which bypasses the need to  compute the integral $\int_{\mathcal{S}} \phi(s, a)^\top \mu(s^\prime)p(s^\prime) ds^\prime$.  
%
%
Specifically, given transition data $\{(s_i,a_i,s'_i)\}_{i=1}^n$, we solve a version of \eqref{eq:ranking} 
where
$x_i=s_i^\prime$, $u_i=(s_i,a_i)$, $f(s^\prime,(s,a))\propto\phi(s, a)^\top p(s^\prime)\mu(s^\prime)$.
As introduced in~\secref{sec:prelim}, NCE can maximize an objective like \eqref{eq:binary} or \eqref{eq:ranking} 
without requiring  computation of the integral $\int_{\mathcal{S}} \phi(s, a) \mu(s^\prime) p(s^\prime) ds^\prime$ during the training stage. We only need a predefined noise distribution $q\rbr{s^\prime}$ that has full support on the domain of interest to generate the negative samples. One can also use $q(s^\prime) = p(s^\prime)$ for simplicity.

\if
by maximizing the objective
\begin{align}
 \sum_{i=1}^n\bigg[ \log(h(x_i, u_i; \gamma)) + \sum_{j=1}^K \log(1-h(y_{i, j}, u_i;  \gamma)) \bigg],
     \label{eq:binary}
\end{align}
over $f\in\mathcal{F}$ and $\gamma\in\mathbb{R}$,
where $h(x, u; \gamma)$ is defined via
\begin{align}
    h(x, u;\gamma) = \frac{f(x, u)\exp(-\gamma)/q(x)}{f(x, u)\exp(-\gamma)/q(x) + K},
\end{align}
\fi
A desirable property of NCE pointed out by \citep{ma2018noise} is that under the realizability assumption with a valid ground truth probability model, the optimal solution will automatically satisfy
\begin{align*}
    \int_{\mathcal{S}} \langle \phi^*(s, a), p(s^\prime)  \mu^*(s^\prime)\rangle = 1.
\end{align*}
For numerical purposes, we can also enforce this marginalization constraint
during the entire training path
and not just the endpoint
by
adding the additional regularizer to the loss
\begin{align*}
    &
    \mathbb{E}_{(s, a)} \left[\left(\log \int_{\mathcal{S}}\phi(s, a)^\top \mu(s^\prime)p(s^\prime)
    ds^\prime\right)^2\right]
    \\
    \approx & 
    \frac{1}{n}\sum_{i=1}^n \left(\log \left(\frac{1}{K}\sum_{j=1}^K \phi(s_i, a_i)^\top\mu\rbr{s^\prime_{i, j}}
    \right)\right)^2,
\end{align*}
where $\{s_{i, j}^\prime\}_{j=1}^K \sim p(s^\prime)$. 

Another important property of NCE is that \eqref{eq:binary} and \eqref{eq:ranking} have a close relationship with MLE, as we will show in~\secref{sec:justification}. This will allow us to show  that the NCE solution preserves statistical properties of MLE that are key to ensuring the  sample complexity of policy optimization. 

Finally, we note that there are other estimation methods for unnormalized  models, such as score matching~\citep{hyvarinen2005estimation}, contrastive divergence~\citep{hinton2002training} and adversarial training~\citep{dai2019exponential}, that can be used for representation learning in a linear MDP. However, for most of these estimators, it is not clear how to establish the estimation error bound in terms of the total variation or Hellinger distances, which breaks a key step in establishing  the needed statistical property for policy optimization. 



\begin{algorithm}[tb] 
\caption{\algabb-LCB: Offline Representation Learning}
\label{alg:offline_algorithm}
\begin{algorithmic}[1]
\State \textbf{Input:} Regularizer $\lambda$, Parameter $\alpha$, Model class $\mathcal{M}$, Dataset $\mathcal{D}$, Number of Negative Samples $K$.

\State Learn representation $\widehat{\phi}(s, a)$ via~\eqref{eq:ranking}. \label{line:representation_offline}  
\State Set the empirical covariance matrix 
\begin{equation*}
    \widehat \Sigma=\sum_{(s,a)\in \Dcal}\widehat \phi(s,a)\widehat \phi(s,a)^\top+\lambda I.
\end{equation*}
\State Set the reward penalty: 
    \begin{equation*}
    \widehat b (s,a)=\min \left(\alpha \sqrt{\widehat \phi(s,a)^{\top}\widehat \Sigma^{-1}\widehat \phi(s,a)},2\right). %
    \end{equation*}\label{line:penalty}
\State Solve \label{line:offline_plan}
$\widehat \pi =\argmax_{\pi}V^{\pi}_{\widehat P,r-\widehat b}.$
\State \textbf{Return } $\hat{\pi}$
\end{algorithmic}
\end{algorithm}

\subsection{\algabb Algorithm Framework}\label{sec:full_alg}
With the newly proposed modifications, we now describe the algorithmic framework for \algabb. Central to \algabb is the learning of the representations $\phi$ and $\mu$ via NCE (Line \ref{line:representation_online} in Algorithm \ref{alg:online_algorithm} and Line \ref{line:representation_offline} in Algorithm \ref{alg:offline_algorithm}.
In the offline setting, we already have a dataset $\mathcal{D}$, hence we can directly perform NCE on $\mathcal{D}$ to obtain the (normalized) representation $\widehat{\phi}$. However, in the online setting, we need to collect a dataset with sufficient quality (Line \ref{line:data_collection}). Following \citet{uehara2021representation}, we first draw a sample from the stationary distribution of the policy proposed in the previous iteration (\ie, $\pi_{n-1}$, which can be done by stopping with probability $1-\gamma$ at each step when  interacting with the environment), then uniformly sample an action to guarantee sufficient exploration. For the optimism and pessimism needed in the online and offline scenarios, it is sufficient to
add the standard deviation as a reward bonus (Line \ref{line:bonus} in Algorithm \ref{alg:online_algorithm}) or substract it as a reward penalty (Line \ref{line:penalty} in Algorithm \ref{alg:offline_algorithm}).
The detailed algorithms are given in Algorithm \ref{alg:online_algorithm} and~\ref{alg:offline_algorithm}.


\paragraph{Planning Module:} 
We need a planning module for the policy optimization components in~Line \ref{line:online_plan} of Algorithm \ref{alg:online_algorithm} and~Line \ref{line:offline_plan} of Algorithm \ref{alg:offline_algorithm}, respectively. 
Here we provide a  practical planning algorithm that exploits the linear structure of the  learned representation $\phi\rbr{s, a}$, rather than adopt a planning module from existing model-based RL approaches~\citep{chua2018deep,kurutach2018model,hafner2019learning}, which only use samples from the learned model and thus are too inefficient. We illustrate optimistic planning in~\algabb-UCB as a concrete example;  a pessimistic planner for~\algabb-LCB is given in~\appref{appendix:planner}.

Given the
representation $\phi\rbr{s, a}$, 
the Bellman equation is 
\begin{equation}\label{eq:bellman_linear_mdp}
Q^\pi\rbr{s, a} = r\rbr{s, a} + \bhat_n\rbr{s, a} + \gamma \EE_{P}\sbr{V^\pi\rbr{s'}},
\end{equation}
which implies that $Q^\pi$  no longer lies in the linear space of $\phi\rbr{s, a}$ due to the bonus. Fortunately, if we augment the space with $\psi\rbr{s, a} = \sbr{\phi\rbr{s, a}, \bhat_n\rbr{s, a}}$, we can still represent the $Q$-function linearly with $\psi\rbr{s, a}$. 
However, directly introducing $\bhat_n\rbr{s, a}$ 
leads to an extra $\Ocal\rbr{d^2}$ memory cost. For practical  considerations, we exploit the fact that $\bhat_n\rbr{s, a}$ is a nonlinear function of $\phi\rbr{s, a}$ and parametrize $Q\rbr{s, a} = w_1^\top \phi\rbr{s, a} + w_2^\top\sigma\rbr{w_3^\top\phi\rbr{s, a}}$, where $\sigma\rbr{\cdot}$ is a nonlinear activation function  used to approximate the $\bhat_n\rbr{s, a}$ feature. 
With this
parametrization of the $Q$ family, 
we conduct fitted $Q$-evaluation~\citep{munos2008finite,antos2007fitted} for policy evaluation, and use entropy-regularized policy gradient~\citep{nachum2017bridging,haarnoja2018soft} to improve the policy.
%
 See~\appref{appendix:planner}
for more details about the practical planning module. 

\subsection{Practical Implementation}
We build our practical online algorithm on top of the actor-critic algorithm with entropy regularizer~\citep{haarnoja2018soft}. The only change we make is stopping the gradient back-propagation from the temporal-difference learning objective to the backbone of the critic function, \ie, freezing $\phi$. We update the backbone network by entirely~\eqref{eq:ranking}. For the planning module, since some of our tasks are continuous, we train a policy network with policy gradient steps. We list two important design choices here:

\begin{itemize}
    \item We choose the negative distribution $q$ in~\eqref{eq:ranking} to be a mixture of $50\%$ samples from the replay buffer and $50\%$ samples from random trajectories.
    \item For computation convenience, instead of random sample actions (Line~\ref{line:data_collection} in Algorithm~\ref{alg:online_algorithm}), we choose a mixture of random actions and on-policy actions, \ie, $\epsilon\pi_0(a|s) + (1-\epsilon)\pi(a|s)$, where $\pi_0(a|s)$ is a uniform policy. Detailed ablations are in~\appref{appendix:mix_sampling}.
\end{itemize}

\section{Theoretical Justification}\label{sec:justification}
We follow the outline of \citep{uehara2021representation} to provide a  theoretical justification for  \algabb in terms of its sample complexity in both the online and offline settings. 
\begin{theorem}[Sample Complexity of Online \algabb-UCB]\label{thm:online_ucb}
Assume $P \in \mathcal{M}$ where $\mathcal{M}$ is a finite model class, $|\mathcal{A}| < \infty$. Let $\alpha_n = O(\sqrt{(|\mathcal{A}| + d^2) \gamma \log (|\mathcal{M}|n/\delta)})$, $\lambda_n = O(d\log (|\mathcal{M}|n/\delta))$. As $K\to\infty$, \algabb-UCB needs at most $\widetilde{O}\left(\frac{d^4|\mathcal{A}|^2}{(1-\gamma)^5\epsilon^2}\right)$ samples to identify a policy $\widehat{\pi}$ such that
\begin{align*}
    V_{P^*, r}^{\pi^*} - V_{P^*, r}^{\widehat{\pi}} \leq \epsilon.
\end{align*}
\end{theorem}
\begin{theorem}[Error Bound of Offline \algabb-LCB]\label{thm:offline_lcb}
Assume $P \in \mathcal{M}$ where $\mathcal{M}$ is a finite model class, $|\mathcal{A}| < \infty$. Define
\begin{align*}
    C_\pi^* = & \mathrm{tr}\left(\mathbb{E}_{d_{P}^\pi}[\phi^*(s, a) \phi^*(s, a)^\top]\right.\\
    & \quad \quad \left.(\mathbb{E}_{\mathcal{D}}[\phi^*(s, a) \phi^*(s, a)^\top])^{-1}\right),
\end{align*}
where $\mathcal{D}$ is the empirical measure of the offline data $\{(s_i, a_i, r_i, s_i^\prime)\}_{i=1}^n$. Let $\omega = \max_{a, s} \pi_b(a|s)$ where $\pi_b$ is the behavior policy. Let $\alpha = \sqrt{(\omega + d^2) \gamma \log |\mathcal{M}|/\delta}$, $\lambda = O(d\log |\mathcal{M}|/\delta)$. As $K\to\infty$, for any policy $\pi$, we have that
\begin{align*}
    V_{P^*, r}^{\pi} - V_{P^*, r}^{\widehat{\pi}} = \tilde{O}\left(\frac{d^2 \omega}{(1-\gamma)^2} \sqrt{\frac{C_\pi^*}{n}}\right).
\end{align*}
\end{theorem}

We use $\widetilde{O}\rbr{\cdot}$ to denote order up to logarithmic factors. We provide a proof sketch, deferring details to~\appref{appendix:proof}. 
%
\begin{proof_sketch}
The main difference between our proposed method and \citep{uehara2021representation} is that we use NCE for the representation learning, while \citet{uehara2021representation} use an MLE oracle, which is computationally intractable. We first establish the consistency of NCE w.r.t.\ MLE:
\begin{lemma}
\label{lem:consistency}
Consider the conditional model $p(x|u)$ defined as $\frac{f(x, u)}{Z_f(u)}$ where $Z_f(u) = \int f(x, u) dx$. Define
\begin{align}
    \widehat{p}_n^{\mathrm{MLE}} = \mathop{\arg\max}_{f\in\mathcal{F}} \frac{1}{n}\sum_{i=1}^n \log {f(x_i, u_i)}/{Z_f(u_i)},
\end{align} 
and  $\widehat{p}_n^{\mathrm{NCE}}(x|u) = \frac{\widehat{f}_n^{\mathrm{NCE}}(x, u)}{Z_{\widehat{f}_n^{\mathrm{NCE}}}(u)}$.
The estimator obtained with the ranking objective \eqref{eq:ranking} satisfies the following property almost surely:
\begin{align}
    \lim_{K\to\infty}\widehat{p}_n^{\mathrm{NCE}, \mathrm{ranking}} =\widehat{p}_n^{\mathrm{MLE}}
\end{align}
Moreover,
if $Z_f(u)$ is a constant function that does not depend on $u$ for all $f\in \mathcal{F}$, 
then the estimator obtained by  \eqref{eq:binary} satisfies the following property almost surely:
\begin{align}
    \lim_{K\to\infty}\widehat{p}_n^{\mathrm{NCE}, \mathrm{binary}} =\widehat{p}_n^{\mathrm{MLE}}
\end{align}
\end{lemma}
With Lemma~\ref{lem:consistency}, we know when $K\to\infty$, NCE and MLE will return the same estimator. Then following the proof of \citep{uehara2021representation}, we can obtain the stated results.
\end{proof_sketch}

We considered the sample complexity with a number of negative samples $K\rightarrow \infty$ in NCE. The non-asymptotic error of NCE in terms of the total variation distance or the Hellinger distance is still an open problem in density estimation. Standard empirical process theory (e.g. \citep[Chapter 14]{sugiyama2012density}) can only show the convergence rate in terms of the objective, but it can be non-trivial to bound the measure of interest with only convergence in the objective. As our purpose is making linear MDPs practical, we leave such theoretical nuances of the original NCE as future work. We also note that, there exists an alternative form of the contrastive loss, which was first introduced in the representation learning \citep{tosh2021contrastive} when the data have some additional structures like the multi-view structure \citep{sridharan2008information, foster2008multi}. The concurrent work~\citep{qiu2022contrastive} provides a non-asymptotic TV error bound of $\mathcal{O}(1/\sqrt{n})$ without the assumption on specific structure. We leave the comparison of the original NCE with this alternative contrastive loss as future work.
\thmref{thm:online_ucb} and~\thmref{thm:offline_lcb} rely on some conditions about the model class size, action space, and coverage assumption, which we discuss in the following.

\vspace{-1em}
\paragraph{Remark (Infinite model class):}
\citet{agarwal2020flambe, uehara2021representation} focused on the case when the model class $\mathcal{M}$ is finite and $P\in\mathcal{M}$, which simplifies the theoretical analysis. However, in practice, since we use a function approximator with sufficient expressive power like a neural network, it is likely that $P\in \mathcal{M}$ more generally.  When satisfied, the theoretical results will still hold for infinite $\mathcal{M}$ with an additional logarithmic complexity term, such as effective dimension \citep{zhang2002effective} or Rademacher complexity \citep{bartlett2002rademacher}.

\vspace{-1em}
\paragraph{Remark (Infinite action space):}
The assumption $|\mathcal{A}| < \infty$ is restrictive. However, there are no known results that remove this dependence for representation learning due to the need for exploration. In principle, we can replace $|\mathcal{A}|$ with the volume of the action set, which would be statistically and computationally manageable provided the dimensionality of the action space was small.

\vspace{-1em}
\paragraph{Remark (Coverage assumption):}
$C_{\pi}^*$ in the offline RL scenarios measures the quality of the offline data under the optimal representation $\phi^{*}$. If the offline data span the optimal representation space well, then the spectrum of $(\mathbb{E}_{\mathcal{D}}[\phi^*(s, a) \phi^*(s, a)^\top])^{-1}$ is small, implying that $C_{\pi}^*$ is small and we  obtain a near optimal policy. 


\section{Experiments}\label{sec:exp}

\begin{table*}[t!]
\caption{\footnotesize Performance on various MuJoCo control tasks. All the results are averaged across 4 random seeds and a window size of 10K. Results marked with $^*$ is adopted from MBBL~\citep{wang2019benchmarking}. \algabb achieves strong performance compared with baselines.}
\scriptsize
\setlength\tabcolsep{3.5pt}
\label{tab:MuJoCo_results}
\centering
\begin{tabular}{p{2cm}p{2.5cm}p{2cm}p{2.5cm}p{2cm}p{2cm}p{2cm}p{2cm}}
\toprule
& & HalfCheetah & Reacher & Humanoid-ET & Pendulum & I-Pendulum \\ 
\midrule  
\multirow{4}{*}{Model-Based RL} & ME-TRPO$^*$ & 2283.7$\pm$900.4 & -13.4$\pm$5.2 & 72.9$\pm$8.9 & \textbf{177.3$\pm$1.9} & -126.2$\pm$86.6\\
& PETS-RS$^*$  & 966.9$\pm$471.6 & -40.1$\pm$6.9 & 109.6$\pm$102.6 & 167.9$\pm$35.8 & -12.1$\pm$25.1\\
& PETS-CEM$^*$  & 2795.3$\pm$879.9 & -12.3$\pm$5.2 & 110.8$\pm$90.1 & 167.4$\pm$53.0 & -20.5$\pm$28.9\\
& Best MBBL & 3639.0$\pm$1135.8 & \textbf{-4.1$\pm$0.1} & 1377.0$\pm$150.4 & \textbf{177.3$\pm$1.9} & \textbf{0.0$\pm$0.0}\\
\midrule
\multirow{3}{*}{Model-Free RL} & PPO$^*$ & 17.2$\pm$84.4 & -17.2$\pm$0.9 & 451.4$\pm$39.1 & 163.4$\pm$8.0 & -40.8$\pm$21.0 \\
& TRPO$^*$ & -12.0$\pm$85.5 & -10.1$\pm$0.6 & 289.8$\pm$5.2 & 166.7$\pm$7.3 & -27.6$\pm$15.8 \\
& SAC$^*$ (3-layer)  & 4000.7$\pm$202.1 & -6.4$\pm$0.5 & \textbf{1794.4$\pm$458.3} & 168.2$\pm$9.5 & -0.2$\pm$0.1\\
\midrule
\multirow{4}{*}{Representation RL} & DeepSF & 4180.4$\pm$113.8 & -16.8$\pm$3.6 & 168.6$\pm$5.1 & 168.6$\pm$5.1 & -0.2$\pm$0.3\\
& SPEDE & 4210.3$\pm$92.6 & -7.2$\pm$1.1 & 886.9$\pm$95.2 & 169.5$\pm$0.6 & 0.0$\pm$0.0\\
& {\bf \ucblinear} & \textbf{4667.4$\pm$341.5} & -7.3$\pm$0.7 & 1165.6$\pm$144.7 & {167.8$\pm$1.3} & \textbf{0.0$\pm$0.0} \\
& {\bf \ucbmlp (2-layer)} & \textbf{4888.5$\pm$205.8} & \textbf{-6.0$\pm$0.2} & \textbf{1605.9$\pm$225.1} & {168.1$\pm$3.1} & \textbf{0.0$\pm$0.0}\\
\bottomrule 
\end{tabular}
\centering
\begin{tabular}{p{2cm}p{2.5cm}p{2cm}p{2.5cm}p{2cm}p{2cm}p{2cm}p{2cm}}
& & Ant-ET & Hopper-ET & S-Humanoid-ET & CartPole & Walker-ET \\ 
\midrule  
\multirow{4}{*}{Model-Based RL} & ME-TRPO$^*$ & 42.6$\pm$21.1 & 1272.5$\pm$500.9 & -154.9$\pm$534.3 & 160.1$\pm$69.1 & -1609.3$\pm$657.5\\
& PETS-RS$^*$ & 130.0$\pm$148.1 &  205.8$\pm$36.5 & 320.7$\pm$182.2 & 195.0$\pm$28.0 & 312.5$\pm$493.4 \\
& PETS-CEM$^*$ & 81.6$\pm$145.8 & 129.3$\pm$36.0 & 355.1$\pm$157.1 & 195.5$\pm$3.0 & 260.2$\pm$536.9 \\
& Best MBBL & 275.4$\pm$309.1 & 1272.5$\pm$500.9 & \textbf{1084.3$\pm$77.0} & \textbf{200.0$\pm$0.0} & 312.5$\pm$493.4\\
\midrule
\multirow{3}{*}{Model-Free RL} & PPO$^*$ & 80.1$\pm$17.3  & 758.0$\pm$62.0 & 454.3$\pm$36.7 & 86.5$\pm$7.8 & 306.1$\pm$17.2\\
& TRPO$^*$ & 116.8$\pm$47.3  & 237.4$\pm$33.5 & 281.3$\pm$10.9 & 47.3$\pm$15.7 & 229.5$\pm$27.1\\
& SAC$^*$ (3-layer) & 2012.7$\pm$571.3  & 1815.5$\pm$655.1 & 834.6$\pm$313.1 & \textbf{199.4$\pm$0.4} & \textbf{2216.4$\pm$678.7}\\
\midrule
\multirow{4}{*}{Representation RL} & DeepSF & 768.1$\pm$44.1  & 548.9$\pm$253.3 & 533.8$\pm$154.9 & 194.5$\pm$5.8 & 165.6$\pm$127.9\\
& SPEDE & 806.2$\pm$60.2  & 732.2$\pm$263.9 & 986.4$\pm$154.7 & 138.2$\pm$39.5 & 501.6$\pm$204.0\\
& {\bf \ucblinear} & \textbf{3032.6$\pm$495.3} & 775.2$\pm$342.7 & \textbf{1178.8$\pm$65.4} & 180.4$\pm$26.8 & 1670.5$\pm$824.2  \\
& {\bf \ucbmlp (2-layer)} & \textbf{1806.8$\pm$1488.0} & \textbf{2910.6$\pm$323.1}  & \textbf{1257.9$\pm$262.8} & \textbf{199.4$\pm$1.3} & \textbf{3241.2$\pm$761.4}\\
\bottomrule 
\end{tabular}
\end{table*}
In this section, we first demonstrate that NCE is capable of capturing environment transition dynamics in a handcrafted environment, and then
show that \algabb-UCB achieves state-of-the-art performance in MuJoCo~\citep{1606.01540} and DeepMind Control Suite~\citep{tassa2018deepmind} benchmarks, compared to both model-based (\eg, PETS~\citep{chua2018deep}, ME-TRPO~\citep{kurutach2018model}) and model-free RL algorithms (\eg, SAC~\citep{haarnoja2018soft}).
We also instantiate an offline version of our algorithm \algabb-LCB with just one line of modification from \algabb-UCB and show that it achieves comparable performance with some SoTA algorithms (e.g., CQL~\citep{kumar2020conservative}). 
To the best of our knowledge, this is the first practical RL algorithm that leverages the linear MDP assumption to surpass current SoTA model-free and model-based RL algorithms,
taking a significant step toward making linear MDP algorithms practical and bridging the gap between theoretical and empirical RL. 
Note that as discussed before, due to the fact that bonus and entropy term does not lie in the linear space of $\phi(s, a)$, we instantiate two version of our algorithm: \ucblinear uses a linear layer for the $Q$-network while \ucbmlp adopts a 2-layer MLP. We adopt 2 or 3 layers of MLP as $Q$-network for baselines (\eg, SAC) as a fair comparison. Please refer to~\appref{appendix:linear_mlp} for the details. We also provide all the learning curves in~\appref{appendix:add_exp}.

\begin{table*}[t]
\caption{\footnotesize Performance of on various Deepmind Control Suite tasks. All the results are averaged across four random seeds and a window size of 10K. Comparing with SAC, our method achieves even better performance on sparse-reward tasks.}
\scriptsize
\setlength\tabcolsep{3.5pt}
\label{tab:DM_results}
\centering
\begin{tabular}{p{2cm}p{2.5cm}p{1.5cm}p{2cm}p{1.5cm}p{2cm}p{1.5cm}p{1.5cm}}
\toprule
& & cheetah\_run & cheetah\_run\_sparse & walker\_run & walker\_run\_sparse & humanoid\_run & hopper\_hop \\ 
\midrule
\multirow{3}{*}{Model-Free RL} & PPO & 227.7$\pm$57.9 & 5.4$\pm$10.8 & 51.6$\pm$1.5 & 0.0$\pm$0.0 & 1.1$\pm$0.0 & 0.7$\pm$0.8\\
& SAC (2-layer)  & 222.2$\pm$41.0 & 32.4$\pm$27.8 & 183.0$\pm$23.4 & 53.5$\pm$69.3 & 1.3$\pm$0.1 & 0.4$\pm$0.5\\
& SAC (3-layer)  & 595.2$\pm$96.0 & 419.5$\pm$73.3 & 700.9$\pm$36.6 & 311.5$\pm$361.4 & 1.2$\pm$0.1 & 28.6$\pm$19.5\\
\midrule  
\multirow{3}{*}{Representation RL} & DeepSF & 295.3$\pm$43.5 & 0.0$\pm$0.0 & 27.9$\pm$2.2 & 0.1$\pm$0.1 & 0.9$\pm$0.1 & 0.3$\pm$0.1 \\
& {\bf \ucbmlp (2-layer)} & \textbf{679.0$\pm$40.8} & \textbf{446.2$\pm$146.3} & \textbf{743.1$\pm$53.0} & \textbf{697.0$\pm$44.8} & \textbf{11.5$\pm$5.4} & \textbf{161.9$\pm$76.1}\\
\bottomrule 
\end{tabular}
\end{table*}

\begin{table*}[h]
\centering
\caption{\footnotesize Average normalized scores on D4RL locomotion tasks. \algabb-LCB show comparable or better performance especially in expert and medium settings.}
\scriptsize
\label{tab:d4rl}
\begin{tabular}{p{4cm}p{1cm}p{1cm}p{1cm}p{1cm}p{1cm}||p{1cm}p{1cm}p{1cm}}
\toprule
Task Name &SAC &BCQ &BEAR &CQL &BRAC &CPC &\algabb-LCB \\\toprule
halfcheetah-medium-v0 &-4.3 &35.4 &38.6 &42.8 &44.4 &44.1 &\textbf{46.3} \\
hopper-medium-v0 &0.8 &37.1 &47.6 &58.0 &37.2 &\textbf{70.8} &49.9 \\
walker2d-medium-v0 &0.9 &19.2 &33.2 &\textbf{79.2} &69.7 &68.3 &75.3 \\\midrule
halfcheetah-medium-expert-v0 &1.8 &59.2 &51.7 &62.4 &24.0 &66.6 &\textbf{99.3} \\
hopper-medium-expert-v0 &1.6 &66.6 &4.0 &\textbf{111.0} &57.6 &\textbf{112.4} &\textbf{111.3} \\
walker2d-medium-expert-v0 &1.9 &19.1 &10.8 &98.7 &0.8 &93.9 &\textbf{104.2} \\\midrule
halfcheetah-medium-replay-v0 &-2.4 &29.5 &36.2 &\textbf{46.2} &44.3 &45.0 &\textbf{47.5} \\
hopper-medium-replay-v0 &3.5 &15.7 &25.3 &\textbf{48.6} &28.6 &28.4 &31.5 \\
walker2d-medium-replay-v0 &1.9 &8.3 &10.8 &\textbf{26.7} &10.1 &1.0 &22.4 \\
\bottomrule
\end{tabular}
\vspace{-1em}
\end{table*}

\subsection{Learning Maze Dynamics with NCE}
\begin{figure}[t]
    \centering
    \includegraphics[width=0.5\textwidth]{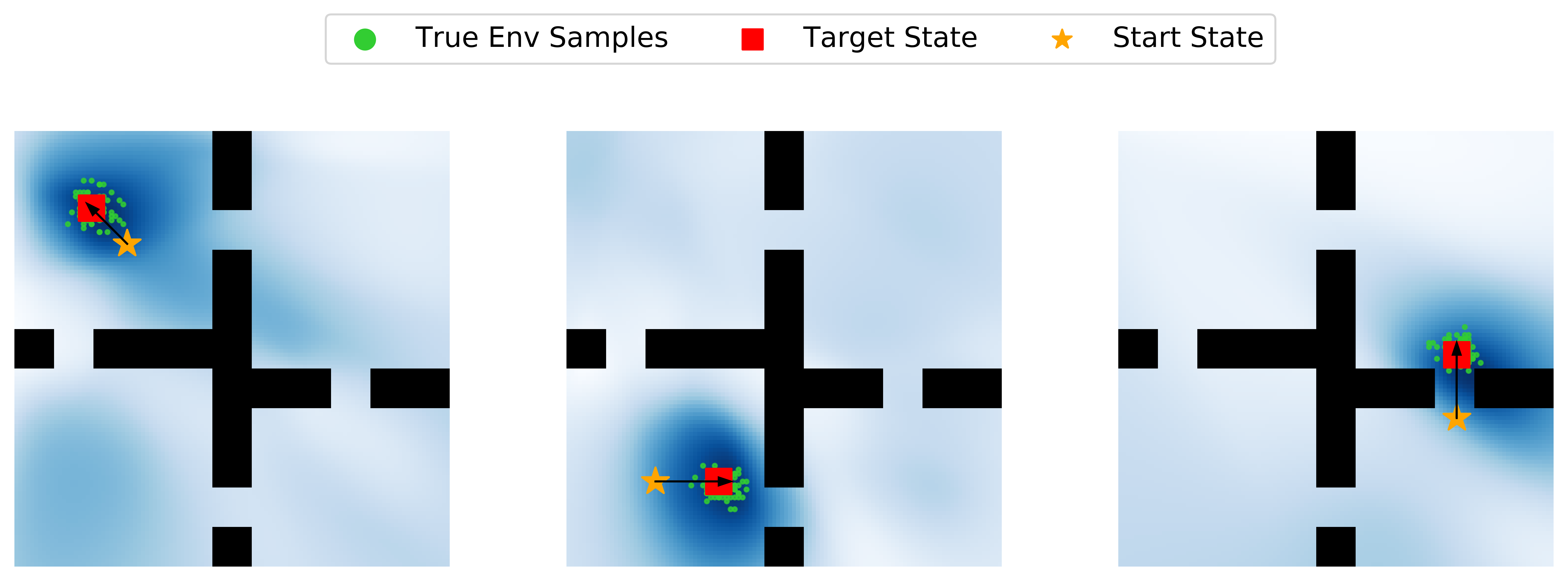}
    \vspace{-2em}
    \caption{\footnotesize \textbf{Heatmap of the Learned Transition:}  The agent transits from orange star to red square and the green dots are the samples drawn from environment simulation.}
    \label{fig:heatmap}
    \vspace{0em}
\end{figure}

We first show that learning with the NCE objective is capable of capturing the transition of the environment. The environment we use here is a four-room maze environment, where the state is the position of the agent and the action is the velocity. The transition can be expressed as $s^\prime = s + a t + \epsilon$, where $t$ is a fixed time interval and $\epsilon \sim \mathcal{N}(0, I)$. We run \ucblinear for $100K$ steps and the learned transition heatmap is visualized in~\figref{fig:heatmap}. 
We see that the high density region is centered around the red square (target position) and and cover the green dots (samples drawn from the environment), which means the representation learned by NCE captures the environment transition and this demonstrates the effectiveness of the representation learning with NCE.

\subsection{Dense-Reward MuJoCo Benchmarks}
We test our algorithm extensively on the dense-reward MuJoCo benchmark from MBBL. We compare \ucblinear and \ucbmlp with model-based RL baselines like PETS, ME-TRPO and model-free RL baselines like SAC and PPO. We also list the best model-based RL results (except for iLQR) in MBBL for comparison. All the algorithms run for $200K$ environment steps. The results are averaged across four random seeds and a window size of $10K$. 

In Table~\ref{tab:MuJoCo_results}, we can see that \ucblinear achieves the SoTA results among all model-based RL algorithms and significantly improves the prior baselines. 
Despite the improvement in terms of final return, prior model-based RL methods can hardly learn the behavior such as walking and moving the ant in these two tasks. By contrast, \ucblinear learns the two behaviors successfully. We also compare our algorithm with the SoTA model-free RL method SAC. Our method achieves comparable sometimes better performance in most of the tasks. Lastly, comparing with two representation learning baselines: DeepSF~\citep{barreto2016successor} and SPEDE~\citep{ren2021free}, \ucblinear also shows dominate performance. However, the performance of \ucblinear is not satisfactory in Humanoid and Hopper environment. It is potentially due to the their hard-exploration nature and the linear parametrized $Q$ failed to capture the bonus for exploration. 

In order to evaluate whether the learned representation is effective for learning the $Q$-function in practice, we also compare \ucbmlp with a 2-layer MLP with the SAC algorithm using a 3-layer MLP for $Q$-network. Our algorithm still achieves impressive performance compared to SAC. This suggests that learning the $Q$-function on top of the learned representation is much more effective.

\vspace{-2mm}
\subsection{Sparse-Reward DeepMind Control Suite}
\label{sec:ablation_bonus}

To better understand how \algabb perform in a sparse-reward setting, we conduct experiments on the DeepMind Control Suite. We compare \ucbmlp with model-free RL methods like SAC (including 2-layer MLP and 3-layer MLP as critic network) and PPO (3 layer of MLP as value network). In addition, we also compare with the representation learning method DeepSF. We run all the algorithms for $200K$ environment steps across four random seeds with a window size of $10K$. \ucbmlp outperforms SAC (2-layer) by a large margin even in relatively dense-reward settings like \texttt{cheeta-run} and \texttt{walker-run}. This margin becomes even larger in sparse-reward setting like \texttt{walker-run-sparse} and \texttt{hopper-hop}. Even comparing with SAC using 3-layer MLP, \ucbmlp also achieves better performance. This again shows the effectiveness of our method.

\begin{table}[h]
\label{tab:ablation_bonus}
\centering
\vspace{-1em}
\caption{\footnotesize Ablation study on bonus coefficient. Bonus boosts the performance especially in sparse-reward settings.}
\scriptsize
\begin{tabular}{p{2.4cm}p{1.3cm}p{1.3cm}p{1.6cm}p{1.5cm}p{2cm}p{1.5cm}p{1.5cm}}
\toprule
& cheetah\_run & walker\_run & walker\_run\_sparse \\ 
\midrule
\ucbmlp ($\alpha=0$) & 656.9$\pm$53.5 & 703.7$\pm$109.0 & 118.4$\pm$236.5\\
\ucbmlp ($\alpha=1$) & 615.5$\pm$28.7 & 700.6$\pm$26.6 & 390.9$\pm$229.7\\
\ucbmlp ($\alpha=5$) & \textbf{679.0$\pm$40.8} & \textbf{743.1$\pm$53.0} & \textbf{697.0$\pm$44.8}\\
\bottomrule
\end{tabular}
\vspace{-1em}
\end{table}

\vspace{-2mm}
\subsection{Ablation Study}
Our ablation study mainly cares about how does the bonus help in online exploration scenarios. To study this, we ablate on the scaling coefficient $\alpha$ used for balancing bonus and actual reward. The results in~\tabref{tab:ablation_bonus} show that in both dense-reward settings like \texttt{cheetah-run} and sparse-reward settings like \texttt{walker-run-sparse}, adding bonus will consistently improve the performance. Especially in \texttt{walker-run-sparse}, the performance boosts from no bonus to $\alpha = 5$ boosts by 583.4. Thus, bonus term is necessary in sparse-reward settings and such optimism help the online learning process. We also show additional ablations on the dimension of feature and ranking vs. classification objective for NCE in Appendix~\ref{appendix:add_exp}.


\vspace{-2mm}
\subsection{Offline Reinforcement Learning}
Lastly, we instantiate our \algabb-LCB algorithm in the offline setting on the D4RL benchmark \citep{fu2020d4rl}. 
Unlike online RL which involves data collection, we split the offline \algabb-LCB into two phases: representation learning phase using NCE and policy update phase. During representation learning phase, we pretrain $\phi(s,a)$ according to Algorithm~\ref{alg:offline_algorithm} with for $200K$ steps. After pretraining, we fix the learned representations and use SAC with moderate policy regularization as in~\citet{wu2019behavior} to learn the downstream tasks for for $1M$ steps across 4 random seeds.

We compare \algabb-LCB with SoTA offline RL algorithms~\citep{fujimoto2019off,kumar2019stabilizing,kumar2020conservative,wu2019behavior} and include CPC~\citep{oord2018representation} as a representation learning baseline. Besides the state-action representation learned in \algabb-LCB, one major difference between CPC and \algabb-LCB is the additional reward penalty introduced by \algabb-LCB. The average normalized scores are presented in~\tabref{tab:d4rl}. \algabb-LCB performs well on the medium-expert set of tasks. The medium set of tasks also benefit from our method. However, \algabb-LCB is unable to improve the performance of medium-replay, potentially due to low-quality data induced representations being incapable of recovering the good policy.

\vspace{-2mm}
\section{Conclusion}
In this paper, we tackle the computational difficulty in linear MDP and make linear MDP practical with SoTA performance, while still preserve the desirable statistical benefits. 

\vspace{-2mm}
\section*{Acknowledgements}
We would like to thank the anonymous reviewers for their detailed and constructive feedback. This project occurred under the Google-BAIR Commons at UC Berkeley. 


\bibliography{references}
\bibliographystyle{icml2022}

\newpage
\appendix
\onecolumn

\section{Details of Proofs}\label{appendix:proof}

\subsection{Consistency of NCE towards MLE}\label{sec:nce}

{\bf Lemma~\ref{lem:consistency}.} {\itshape 
Consider the conditional model $p(x|u)$ defined as $\frac{f(x, u)}{Z_f(u)}$ where $Z_f(u) = \int f(x, u) dx$. Define
\begin{align}
    \widehat{p}_n^{\mathrm{MLE}} = \mathop{\arg\max}_{f\in\mathcal{F}} \frac{1}{n}\sum_{i=1}^n \log {f(x_i, u_i)}/{Z_f(u_i)},
\end{align} 
and  $\widehat{p}_n^{\mathrm{NCE}}(x|u) = \frac{\widehat{f}_n^{\mathrm{NCE}}(x, u)}{Z_{\widehat{f}_n^{\mathrm{NCE}}}(u)}$.
The estimator obtained with the ranking objective \eqref{eq:ranking} satisfies 
\begin{align}
    \lim_{K\to\infty}\widehat{p}_n^{\mathrm{NCE}, \mathrm{ranking}} =\widehat{p}_n^{\mathrm{MLE}}
\end{align}
almost surely.
Moreover,
if $Z_f(u)$ is a constant function that does not depend on $u$ for all $f\in \mathcal{F}$, 
then the estimator obtained by  \eqref{eq:binary} satisfies
\begin{align}
    \lim_{K\to\infty}\widehat{p}_n^{\mathrm{NCE}, \mathrm{binary}} =\widehat{p}_n^{\mathrm{MLE}}
\end{align}
almost surely.
}

In the following part, we will show the proof for the binary objective and ranking objective accordingly.
\paragraph{Binary Objective} The consistency of the binary objective has been discussed in \citep{riou2018noise} under general regularity conditions. We include such result for completeness under more practical assumptions. 
\begin{proof}
First, recall that
\begin{align}
    \left(\widehat{f}_n^{\mathrm{NCE}}, \widehat{\gamma}_n^{\mathrm{NCE}}\right) =& \mathop{\arg\max}_{f\in\mathcal{\mathcal{F}}, \gamma\in\mathbb{R}} \frac{1}{n}  \sum_{i=1}^n\bigg[ \log(h(x_i, u_i; \gamma)) + \sum_{j=1}^K \log(1-h(y_{i, j}, u_i;  \gamma)) \bigg],
      \label{eq:binary2}
\end{align}
where $h(x, u; \gamma)$ is defined via
\begin{align}
    h(x, u;\gamma) = \frac{f(x, u)\exp(-\gamma)/q(x)}{f(x, u)\exp(-\gamma)/q(x) + K}.
\end{align}
Notice that
\begin{align*}
    & \lim_{K\to\infty} \sum_{j=1}^K \log (1 - h(y_{i, j}, u_i; \gamma)) =  -\lim_{K\to\infty} \frac{\sum_{j=1}^K\log\left(1 + \frac{f(y_{i, j}, u_i) \exp(-\gamma)}{K q(y_{i, j})}\right)^{K}}{K} \\
    =  & -\lim_{K\to\infty} \frac{\sum_{j=1}^K \left(\frac{f(y_{i, j}, u_i)\exp(-\gamma)}{q(y_{i, j})}\right)}{K} = -\mathbb{E}_{y_i \sim q(y)}\left[\frac{f(y_{i}, u_i)\exp(-\gamma)}{q(y_{i})}\right] =  -\exp(-\gamma) \int f(y,u_i) d y.
\end{align*}
Meanwhile, we have
\begin{align*}
    & \lim_{K\to\infty} \left[\log(h(x_i, u_i;\gamma)) + \log K\right] = \log \frac{f(x_i, u_i)}{q(x_i)} - \gamma - \lim_{K\to\infty}\log\left(1 + \frac{f(y_{i}, u_i)\exp(-\gamma)}{K q(y_{i}) }\right)\\
    = & \log \frac{f(x_i, u_i)}{q(x_i)} - \gamma.
\end{align*}
Hence, 
\begin{align*}
    & \lim_{K\to\infty}\frac{1}{n}  \sum_{i=1}^n\bigg[ \log(h(x_i, u_i;  \gamma))
    + \sum_{j=1}^K \log(1-h(y_{i, j}, u_i;  \gamma)) \bigg] + \log K =  \frac{1}{n} \sum_{i=1}^n \log \frac{f(x_i, u_i)}{q(x_i)} - \gamma - \exp(-\gamma) \int f(y, u_i) d y.
\end{align*}
Take the derivative w.r.t $\gamma$, we know the maximum is achieved when $\widehat{\gamma}_n^{\mathrm{NCE}} = \log \int f(y, u_i) d y$. As $q$ is a fixed noise distribution, when $K\to\infty$, the maximizer $\widehat{f}_n^{\mathrm{NCE}}$ should also maximize $\frac{1}{n} \log f(x_i, u_i) - \log \int f(y, u_i) d y$, which means it's also the solution of MLE, hence finishes the proof.
\end{proof}
\paragraph{Ranking Objective}
We then show the consistency of the ranking objective.
\begin{proof}
Notice that, with the law of large numbers and continuous mapping theorem, we have
\begin{align*}
    & \lim_{K\to\infty} \log\left(\frac{1}{K}\left(f(x_i, u_i)/q(x_i) + \sum_{j=1}^K f(y_{i, j}, u_i)/q(y_{i, j})\right)\right) = \lim_{K\to\infty}\log\left(\frac{1}{K}\left(\sum_{j=1}^K f(y_{i, j},u_i)/q(y_{i, j})\right)\right)\\
    = & \log \mathbb{E}_{y\sim q(y)} f(y, u_i)/q(y) =  \log \int p(y|u_i) d y.
\end{align*}
Hence, we have
\begin{align*}
    &\lim_{K\to\infty}\frac{1}{n}\sum_{i=1}^n \log \frac{f(x_i, u_i)/q(x_i)}{f(x_i, u_i)/q(x_i) + \sum_{j=1}^K f(y_{i, j}, u_i)/q(y_{i, j})} + \log K = \frac{1}{n} \sum_{i=1}^n \left[\log f(x_i, u_i) / q(x_i) - \log \int f(y, u_i) dy\right].
\end{align*}
With the similar reasoning, we can conclude the proof for the ranking objective.
\end{proof}
\begin{remark}
Notice that, the ranking objective does not require explicit parameter $\gamma$ to approximate the normalizing constant. Hence, for the conditional model $p(y|x)$ where the normalizing constant $\int p(y|x) dy$ can depend on the conditional variable $x$, the ranking objective can still provide consistent estimation, while the binary objective cannot provide consistent estimation, as $\gamma$ is only a scalar that does not depend on $x$ by definition.
\end{remark}

\section{Planning Module Implementation}\label{appendix:planner}


As introduced in~\secref{sec:full_alg}, with the learned $\phi\rbr{s, a}$ fixed, we parametrize the $Q$-function with 
\begin{equation}\label{eq:q_param}
Q_W\rbr{s, a} = w_1^\top \phi\rbr{s, a} + w_2^\top\sigma\rbr{w_3^\top \phi\rbr{s, a}},
\end{equation}
where $W = \rbr{w_1, w_2, w_3}$ denotes the parameters to be learned. We consider policy $\pi_v\rbr{a|s}$, where $v$ denotes the policy parameters. 

For practical consideration, we also introduce entropy-regularization in MDP, which will lead to an extra $\log \pi\rbr{a|s}$ in the Bellman equation~\citep{haarnoja2018soft,nachum2017bridging,dai2018sbeed}, \ie, 
\begin{equation}
Q^\pi\rbr{s, a} = r\rbr{s, a} + \bhat_n\rbr{s, a} + \gamma\EE_{P^\pi}\sbr{Q^\pi\rbr{s', a'}} -\log \pi\rbr{a|s}.
\end{equation}
In fact, for function completeness, we should and one more nonlinear component for capturing $\log \pi\rbr{a|s}$ in~\eqref{eq:q_param}. Empirically, we noticed that the nonlinear part in parametrization~\eqref{eq:q_param} is enough. 

We apply fitted Q-Evaluation with target network, which optimizes the loss functions for $Q$-function in $t$-th iteration, 
\begin{equation}
\ell\rbr{Q_W; \Dcal} = \EE_{\Dcal}\sbr{\rbr{Q_W\rbr{s, a} - \ytil_{s, a} + \log \pi\rbr{a|s}}^2}
\end{equation}
where $\ytil_{s, a} \defeq r\rbr{s, a} + \bhat_n\rbr{s, a} + \Qtil \rbr{s', a'}$, with $\Qtil$ denoting the target network. Then, the policy is updated by policy gradient w.r.t. $v$ as $\EE_\Dcal\sbr{\nabla_v\log \pi_v\rbr{a|s}Q_W(s, a)}$ in online case.

For the offline RL case, we keep everything the same for learning $Q$ except using $-\bhat_n\rbr{s, a}$ as bonus, and we use BRAC~\citep{wu2019behavior} for policy update wihh the learned $Q$. 

We emphasize that we only learn the parameters $W$ in $Q$, and the $\phi\rbr{s, a}$ is fixed without back-propagation in the training stage. This is significantly different from the vanilla model-free $Q$-learning algorithms. In fact, this can largely alleviate the sample complexity since the size of function space is reduced effectively.  

Meanwhile, comparing to the vanilla model-based algorithms, where the learned model is used as simulator for generating samples for optimal policy seeking, the proposed planner extracts the representation function $\phi\rbr{s, a}$ rather than samples only. This exploits more information from the learned model directly, accelerating the policy learning.

\section{Linear and MLP Representations of \algabb-UCB.}\label{appendix:linear_mlp}
As we mentioned in the Section~\ref{sec:method}, the bonus term and entropy term doesn't necessarily lie in the linear space of the representation thus using a linear layer on top of $\phi$ might not be sufficient. To deal with this issue, we build two versions of \algabb-UCB: \ucblinear, which adds an additional loss to try to predict bonus from $\phi(s, a)$ using a linear layer; and \ucbmlp, which builds a 2-layer MLP for predicting the $Q$-value from $\phi(s, a)$. We study both algorithms in the MuJoCo setting and report their performance separately. To distinguish from the baseline SAC and illustrate the effectiveness of representation learning, in DeepMind Control Suite, we use the same neural network architecture for both SAC and \ucbmlp (both use a 2-layer MLP). The detailed network architecture can be found in Appendix.~\ref{appendix:exp_detail}. Therefore, the only difference is that the $Q$-function SAC takes raw state-action pairs as input while \algabb-UCB takes $\phi(s, a)$. We show that \ucblinear has  performance comparable to SAC, while \ucbmlp significantly outperforms SAC using both a 2-layer and a 3-layer MLP for the $Q$-function.

\section{Environment Visualization}\label{appendix:env_vis}
\begin{figure}[t]
    \centering
    \includegraphics[width=0.7\textwidth]{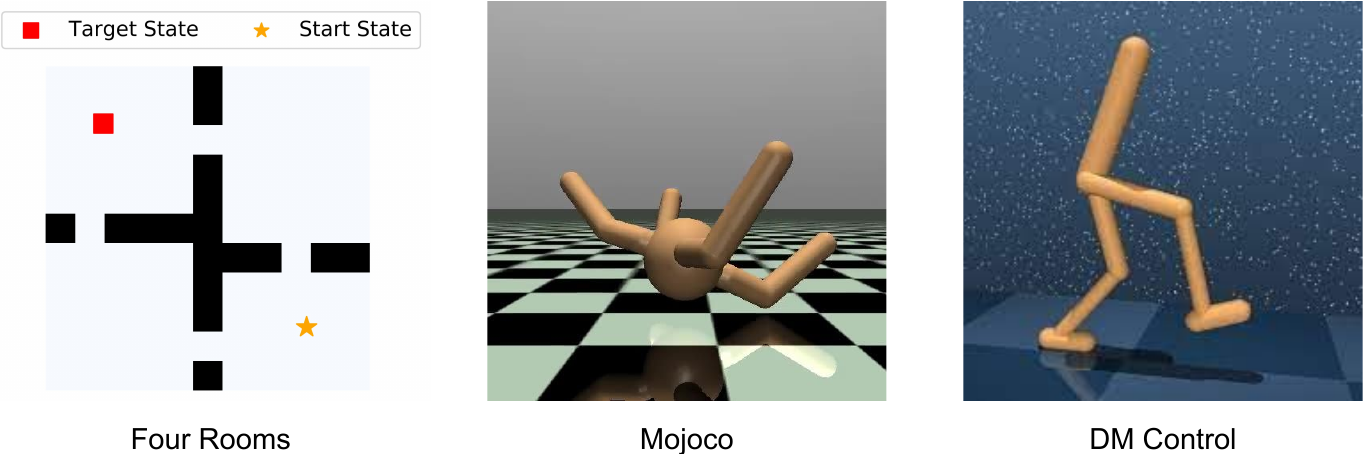}
    \caption{\footnotesize \textbf{Environments:} Visualization all the environments we used in our experiments. (\textit{left}) Hand-designed four-room maze environment. (\textit{middle}) Standard MuJoCo Ant environment. (\textit{right}) DM Control Suite Walker environment.}
    \label{fig:envs}
\end{figure}

We visualize all the environments used for our experiments for a better understanding. In Figure~\ref{fig:envs}, the left one is the hand-designed four-room maze environment. We need to move the agent from the bottom right start state to the upper left target state. Walls shown in black will block all the moves for the agent. The middle one is the Ant environment used in MuJoCo experiments. The task for this is to control the ant so it can crawl. The right one is the Walker task in the DeepMind Control Suite. We need to control its joints and let it walk/run fast. 
\newpage

\section{Experiment Details}\label{appendix:exp_detail}
In this section, we list all the hyperparameter and network architecture we use for our experiments. For online MuJoCo and DM Control tasks, the hyperparameters can be found at Table~\ref{tab:hyper_online}. For bonus scaling coefficient in MuJoCo, we found that since the environment has super dense reward and it is well manually designed, we don't need any bonus for exploration. Therefore, we set this term to 0 for MuJoCo tasks. However, this bonus is critical to the success of DM Control Suite (especially sparse reward environments) and we put a thorough ablation in Section ~\ref{sec:ablation_bonus}. Note that we use exactly the same actor and critic network architecture for all the algorithms in the DM Control Suite experiment.

\begin{table*}[h]
\caption{Hyperparameters used for \algabb-UCB in all the environments in MuJoCo and DM Control Suite.}
\footnotesize
\setlength\tabcolsep{3.5pt}
\label{tab:hyper_online}
\centering
\begin{tabular}{p{6cm}p{3cm}p{5cm}p{2.5cm}p{2.5cm}p{2cm}p{2cm}}
\toprule
& Hyperparameter Value \\ 
\midrule
NCE temperature & 0.2 \\
Bonus Coefficient (MuJoCo) & 0.0 \\
Bonus Coefficient (DM Control) & 5.0 \\
Actor lr & 0.0003 \\
Model lr & 0.0003 \\
Actor Network Size (MuJoCo) & (256, 256) \\
Actor Network Size (DM Control) & (1024, 1024) \\
NCE Embedding Network Size (MuJoCo) & (1024, 1024, 2048) \\
NCE Embedding Network Size (DM Control) & (1024, 1024, 1024) \\
Critic Network Size (MuJoCo) & (2048, 1) \\
Critic Network Size (DM Control) & (1024, 1) \\
Discount & 0.99\\
Target Update Tau & 0.005 \\
Model Update Tau & 0.005 \\
Batch Size & 256 \\
\bottomrule 
\end{tabular}
\end{table*}

The hyperparameters for D4RL benchmark is listed in Table~\ref{tab:hyper_offline}.

\begin{table*}[h]
\caption{Hyperparameters used for \algabb-LCB in D4RL evaluation.}
\footnotesize
\setlength\tabcolsep{3.5pt}
\label{tab:hyper_offline}
\centering
\begin{tabular}{p{6cm}p{3cm}p{5cm}p{2.5cm}p{2.5cm}p{2cm}p{2cm}}
\toprule
& Hyperparameter Value \\ 
\midrule
NCE temperature & 0.2 \\
Bonus Coefficient & 1.0 \\
Actor lr & 0.0003 \\
Critic lr & 0.001 \\
Representation lr & 0.0003 \\
Behavior regularizer & 1. \\
Actor Network Size & (256, 256) \\
NCE Embedding Network Size & (512, 512, 512) \\
Critic Network Size & (512, 1) \\
Discount & 0.99\\
Target Update Tau & 0.005 \\
Batch Size & 256 \\
\bottomrule 
\end{tabular}
\end{table*}
\newpage
\section{Additional Experiments}\label{appendix:add_exp}
In this section, we provide two additional ablation studies on the feature dimension and classification vs. ranking based NCE objective. We also provide the learning curves for all the ablations and DM Control suite experiments. Note that we didn't compare with SPEDE in this setting since the original paper only contains MuJoCo experiments.

\subsection{Additional Ablations}
In addition to the bonus scaling factor, we are also interested in other two components of \algabb: \textit{(1)} Whether there is a preference for using classification-based NCE versus ranking-based NCE? \textit{(2)} Whether there is a significant effect from the dimension of representation $\phi$? We study these two component seprately in Table~\ref{tab:ablation_feature} and Table~\ref{tab:ablation_classify}.

According to the table, We found that larger size of feature dimension in general helps the performance. Interestingly, ranking based objective yields a slightly more stable performance than the classification based objective. 

\begin{table}[h]
\centering
\caption{Ablation study on feature dimension. Larger feature dimension will help the performance in all the tasks. And sometimes it is critical to the success of the algorithm.}
\label{tab:ablation_feature}
\footnotesize
\begin{tabular}{p{3.4cm}p{1.4cm}p{1.4cm}p{1.4cm}p{1.5cm}p{2cm}p{1.5cm}p{1.5cm}}
\toprule
& cheetah\_run & walker\_run & hopper\_hop \\ 
\midrule
\ucbmlp ($h = 256$) & 471.4$\pm$85.7 & 661.2$\pm$40.7 & 99.1$\pm$84.8\\
\ucbmlp ($h = 512$) & 580.1$\pm$50.9 & 772.5$\pm$17.3 & 116.4$\pm$62.4\\
\ucbmlp ($h = 1024$) & \textbf{679.0$\pm$40.8} & \textbf{743.1$\pm$53.0} & \textbf{161.9$\pm$76.1}\\
\bottomrule 
\end{tabular}
\end{table}

\begin{table}[h]
\centering
\caption{Ablation study on classification-based and ranking-based NCE objective. We found that in general ranking based objective yields a more stable performance. }
\label{tab:ablation_classify}
\footnotesize
\begin{tabular}{p{3.4cm}p{1.4cm}p{1.4cm}p{1.4cm}p{1.5cm}p{2cm}p{1.5cm}p{1.5cm}}
\toprule
& cheetah\_run & walker\_run & hopper\_hop \\ 
\midrule
\ucbmlp (classify) & 455.2$\pm$2.7 & 630.5$\pm$39.1 & 62.6$\pm$34.2\\
\ucbmlp (rank) &\textbf{679.0$\pm$40.8} & \textbf{743.1$\pm$53.0} & \textbf{161.9$\pm$76.1}\\
\bottomrule 
\end{tabular}
\end{table}

\subsection{Policy Sampling with Randomness.}\label{appendix:mix_sampling}
One difference between our method and ~\citep{uehara2021representation} is that we don't uniformly sampling from the policy at each time step to collect data for the representation learning objective. We use mixture of learned policy and uniform policy for action sampling. We here include an ablation on the effect for uniform sampling part in Tab.~\ref{tab:sampling oracle}. We can see the uniform action sampling indeed helps in some environments, 
but the improvement is not consistent across different tasks. We conjecture the entropy-regularized policy may also introduce randomness for uniform action sampling.
\vspace{-4mm}
\begin{table}[h]
\centering
\caption{Ablation study on the sampling protocol. At $t$-th epoch, the action is sampled from $(1-\epsilon)\pi(a|s) + \epsilon\pi_0(a)$. }
\label{tab:sampling oracle}
\footnotesize
\begin{tabular}{p{3.4cm}p{1.6cm}p{1.6cm}p{1.6cm}p{2.4cm}p{2cm}p{1.5cm}p{1.5cm}}
\toprule
& cheetah\_run & walker\_run & walker\_run\_sparse \\ 
\midrule
\ucbmlp ($\epsilon = 0$) & \textbf{679.0$\pm$40.8} & 743.1$\pm$53.0 & 697.0$\pm$44.8\\
\ucbmlp ($\epsilon = 0.01$) & 614.6$\pm$57.2 & 742.3$\pm$22.9 & \textbf{734.9$\pm$24.3}\\
\ucbmlp ($\epsilon = 0.05$) & 632.0$\pm$65.7 & \textbf{777.5$\pm$15.7} & 662.2$\pm$166.3\\
\bottomrule 
\end{tabular}
\end{table}
\vspace{-2mm}

\subsection{Performance Curves for DM Control}
\begin{figure}[h]
    \centering
    \includegraphics[width=0.9\textwidth]{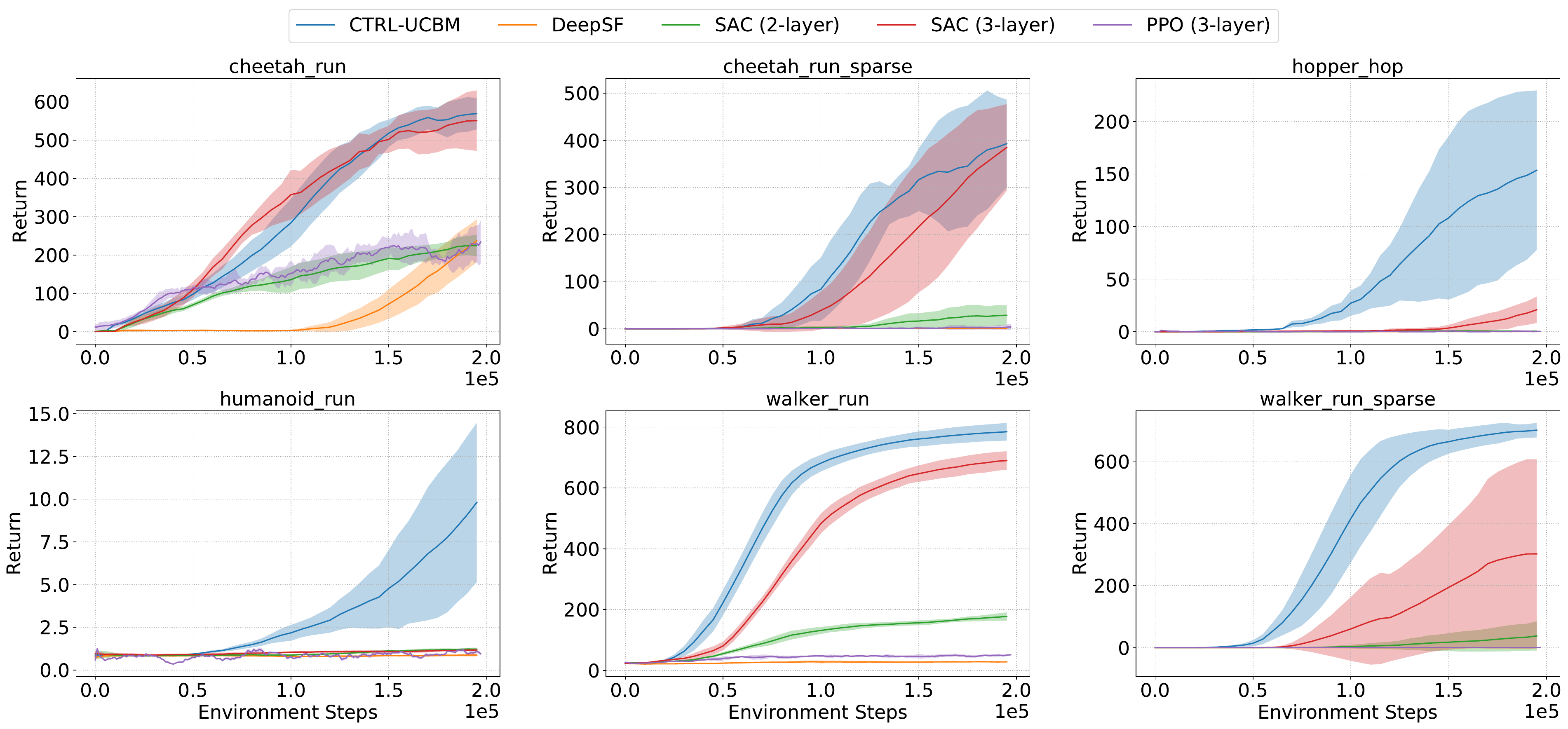}
    \caption{\footnotesize \textbf{Performance for the DM Control Suite Experiments:} We plot the performance curve for DM Control Suite experiments. We can see that \ucbmlp achievs the best performance in all the tasks, even comparing to SAC with 3-layer MLP. }
    \label{fig:appendix_dm_result}
\end{figure}

\begin{figure}[h]
    \centering
    \includegraphics[width=0.9\textwidth]{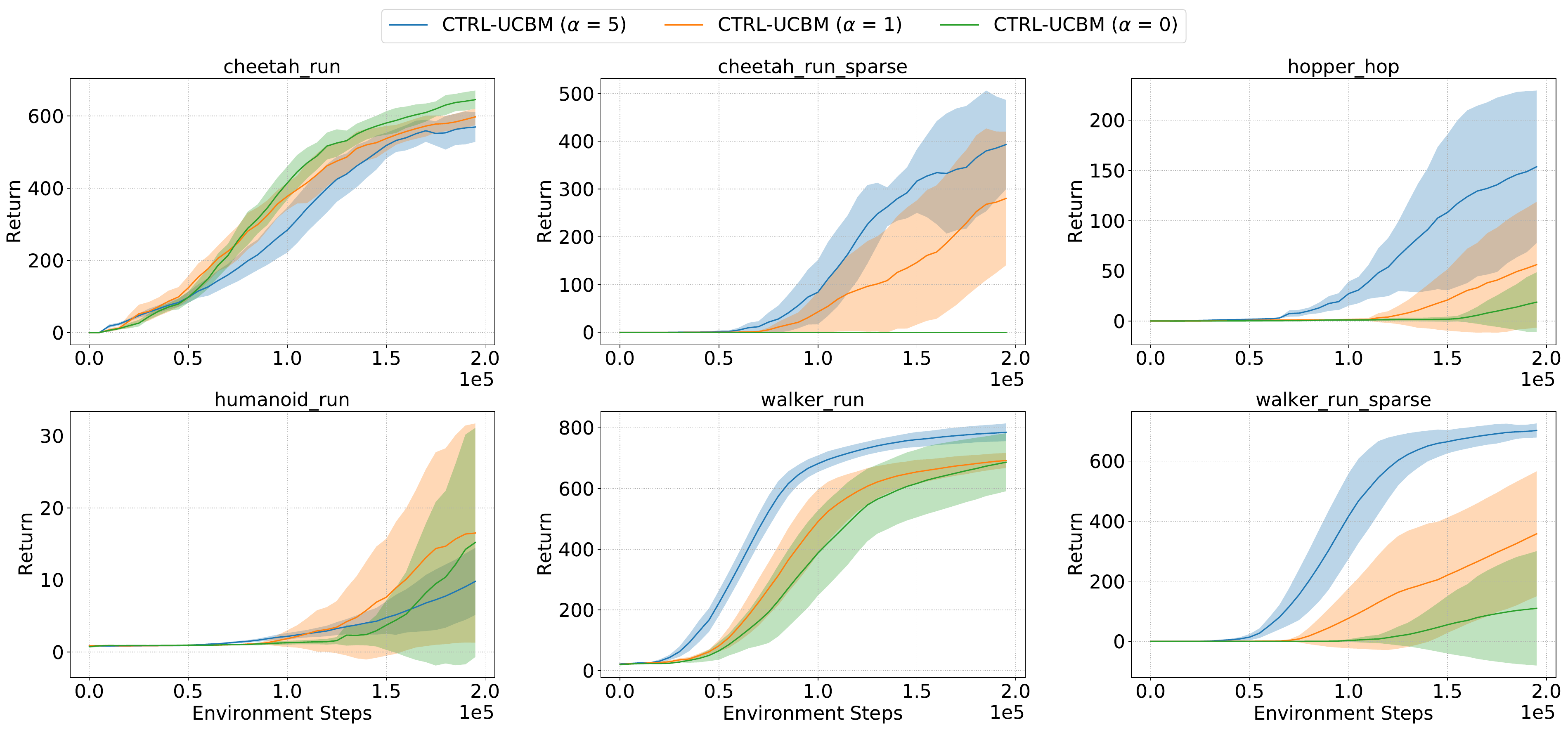}
    \caption{\footnotesize \textbf{Performance Curve for Bonus Scaling Coefficient:} We study the effect of bonus with chaning the bonus scaling factor. We see that with 0 bonus, the algorithm hardly learns in \texttt{cheetah-run-sparse} and \texttt{walker-run-sparse}. This shows the effectiveness of bonus in online exploration settings.}
    \label{fig:appendix_ablation_bonus}
\end{figure}

\begin{figure}[h]
    \centering
    \includegraphics[width=0.9\textwidth]{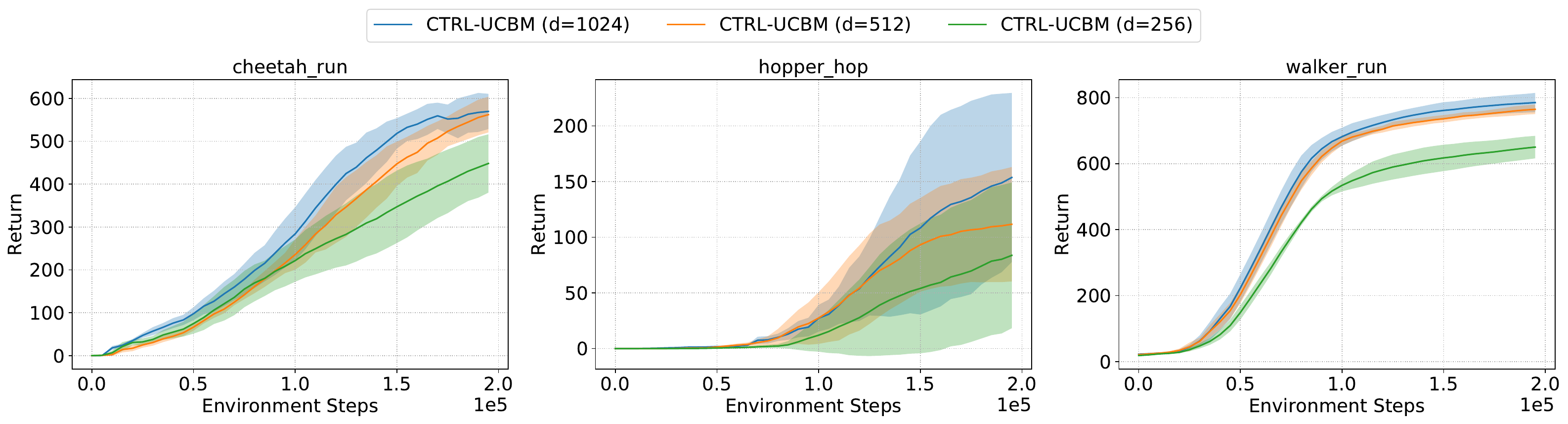}
    \caption{\footnotesize \textbf{Performance Curve for Varying Feature Dimension:} We study the effect of feature dimension in the linear MDP settings. In general the trend is pretty clear: larger feature dimension induces better performance.}
    \label{fig:appendix_ablation_dimension}
\end{figure}

We provide all the performance curves for DM Control experiments. In Figure~\ref{fig:appendix_dm_result}, we clearly see that \ucbmlp surpasses all the baseline methods and the gain is significantly in sparse-reward settings (e.g., in \texttt{walker-run-sparse} and \texttt{hopper-hop}, \ucbmlp surpasses SAC with a good margin). We also provide all the performance curves for ablation studies. In Figure~\ref{fig:appendix_ablation_bonus}, we see that the bonus is critical to the success of the algorithm in sparse-reward settings. In Figure~\ref{fig:appendix_ablation_dimension}, we can conclude that larger feature dimension induces better performance. In Figure~\ref{fig:appendix_ablation_nce}, we can see that ranking-based objective has slightly better performance than classification-based objective. 

\begin{figure}[h]
    \centering
    \includegraphics[width=0.9\textwidth]{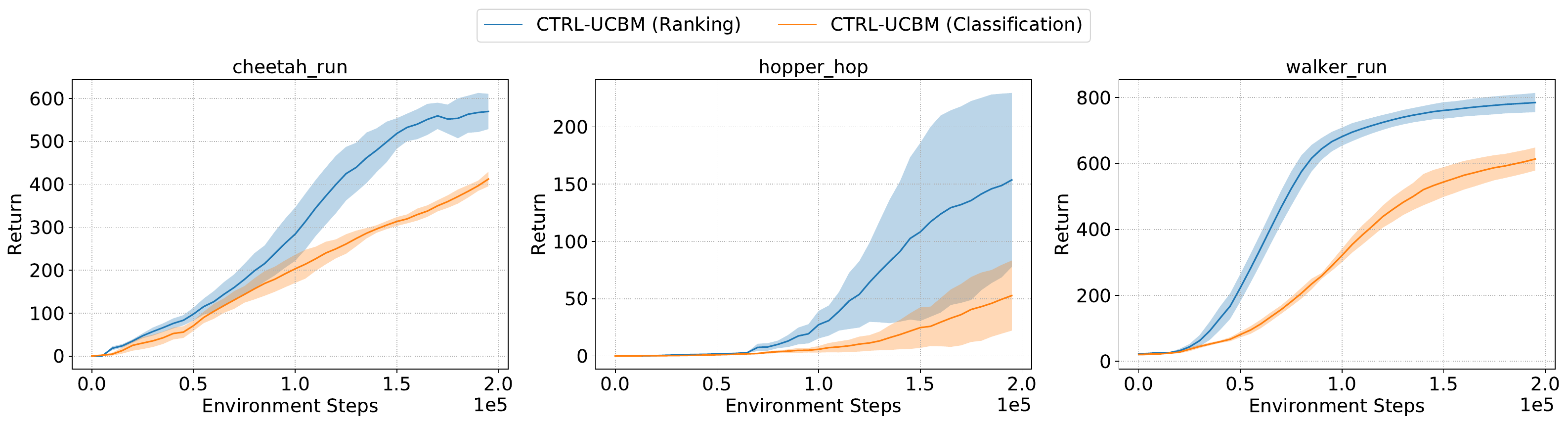}
    \caption{\footnotesize \textbf{Performance Curve for Ranking-Based Objective versus Classification-Based Objective:} We study the effect of using a ranking-based objective for NCE versus classification-based objective for NCE. Interestingly, ranking-based objective results in a slightly stabler performance.}
    \label{fig:appendix_ablation_nce}
\end{figure}

\subsection{Performance Curves for D4RL Benchmarks.}
We also provide learning curves for all the algorithms in D4RL benchmarks. The benefit of \algabb-LCB is most noticable in medium-expert tasks.
\begin{figure}[h]
    \centering
    \includegraphics[width=0.9\textwidth]{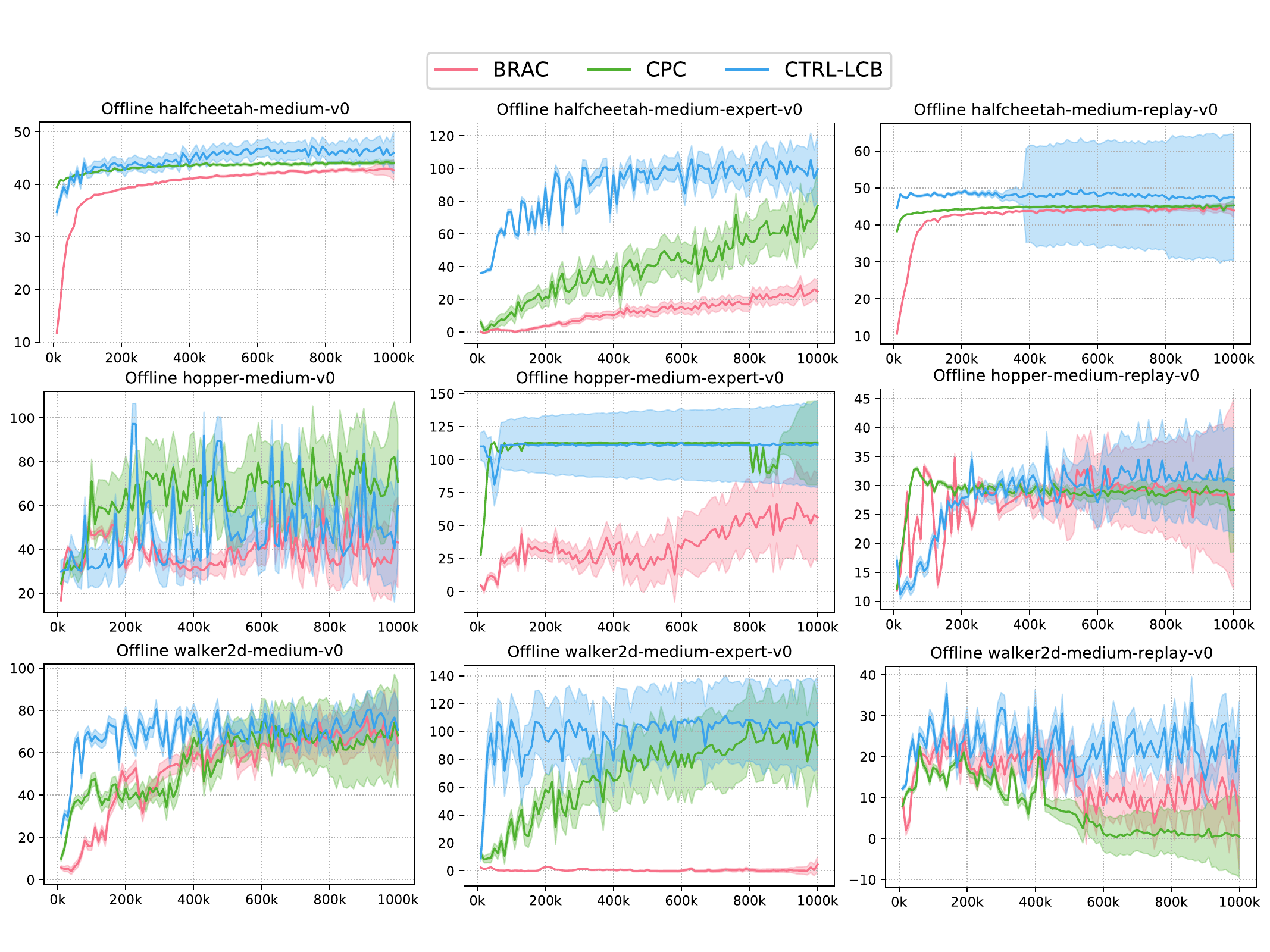}
    \caption{\footnotesize \textbf{Performance Curve for \algabb-LCB on D4RL:} The benefit of \algabb-LCB is the most evident on medium-expert tasks, when the dataset covers some amount of expert demonstration.}
    \label{fig:appendix_ablation_nce}
\end{figure}


\end{document}